# Learning from a Generative AI Predecessor

## The Many Motivations for Interacting with Conversational Agents

**Donald Brinkman**
Mojang Studios
Microsoft Corporation
Redmond, WA, USA
donaldbr@microsoft.com

**Jonathan Grudin**
Information School
University of Washington
Seattle, WA, USA
grudin@uw.edu

## ABSTRACT

For generative AI to succeed, how engaging a conversationalist must it be? For almost sixty years, some conversational agents have responded to any question or comment to keep a conversation going. In recent years, several utilized machine learning or sophisticated language processing, such as Tay, Xiaoice, Zo, Hugging Face, Kuki, and Replika. Unlike generative AI, they focused on engagement, not expertise. Millions of people were motivated to engage with them. What were the attractions? Will generative AI do better if it is equally engaging, or should it be less engaging? Prior to the emergence of generative AI, we conducted a large-scale quantitative and qualitative analysis to learn what motivated millions of people to engage with one such 'virtual companion,' Microsoft's Zo. We examined the complete chat logs of 2000 anonymized people. We identified over a dozen motivations that people had for interacting with this software. Designers learned different ways to increase engagement. Generative conversational AI does not yet have a clear revenue model to address its high cost. It might benefit from being more engaging, even as it supports productivity and creativity. Our study and analysis point to opportunities and challenges.

## 1 Introduction

Fiction writers have long portrayed artificial companions. Mary Shelley's *Frankenstein* was inspired by the harnessing of electricity. The concept moved from science fiction to science in 1949 when Alan Turing wrote, "I do not see why [the computer] should not enter any one of the fields normally covered by the human intellect, and eventually compete on equal terms." [39] The "Turing test" was proposed: programs would converse, by typing, with qualified judges who tried to determine which contestants were people. This inspired engineers, whose work inspired fiction writers to imagine future conversational agents that may be approaching.

In the summer of 1956, mathematicians and engineers gathered and created the field of Artificial Intelligence. They saw intelligence in terms of orderly problem-solving. They underestimated the complexity of human intelligence, concluding that by 1980 or 1985, a computer would achieve human-level intelligence, now called artificial general intelligence. In the words of MIT's Marvin Minsky, it would "be able to read Shakespeare, grease a car, play office politics, tell a joke, have a fight." [6, 12] This lofty goal attracted government funding of the extraordinarily expensive computers of the 1950s and 1960s.

Some people consider generative AI based on large language models (LLMs) to be a significant step toward artificial general intelligence. This article presents a detailed study of a generative AI predecessor, Zo, with which over a million people engaged in extended conversations. Zo relied on machine learning (ML) and sophisticated natural language processing (NLP), but not a large language model. We briefly reviewing decades of efforts to "pass the Turing test." We outline successes, attracting millions of users, and challenges—few conversational agents lasted long. Our study explores why people engaged with Zo and how the development team responded to user behaviors.

| Type | Focus | Sessions | Examples |
|---|---|---|---|
| **Intelligent assistants** | Broad, shallow | 1-3 exchanges | Siri, Cortana, Alexa, Google Assistant, Bixby |
| **Task-focused chatbots** | Narrow, deep | 3-7 exchanges | Dom the Dominos Pizza Bot, customer service & FAQ bots, non-player characters |
| **Virtual companions** | Broad, deep | Unlimited exchanges | ELIZA, ALICE, Cleverbot, Tay, Kuki, Xiaoice, Zo, Hugging Face, Replika, ChatGPT, Bard, Bing AI Chat |

Table 1. Types of conversational agents

Table 1 categorizes conversational agents. Intelligent virtual assistants such as Siri and Alexa respond to short, simple requests. Task-focused agents such as customer-service chatbots have a narrow range and try to keep discussions short. We will focus on the third category, agents that are designed to participate in long discussions. To emphasize this distinction, they are labeled 'virtual companions.'

Virtual companion efforts began in 1965 and picked up momentum around 1990, when Turing test competitions gained prominence. In the late 2010s, a serious effort emerged, driven by more powerful computers, greater storage capacity, advances in machine learning and natural language processing, and commercial expectations that were triggered by intelligent assistants and task-focused chatbots. Kuki, Xiaoice, Replica, Zo, and others gained millions of users. Some are still in use. If you haven't heard of them, it is all the more reason to look closely: Discover why millions of people used them and consider why their use did not spread further.

We set out to understand the motivations of people who engage in long conversations with software. Although our software did not have key capabilities of generative AI, it met needs of millions of people. Generative AI can take note. Generative AI that extends its appeal could drive higher engagement and use. Equally important, challenges in deploying the "virtual companion" Zo that surprised us could complicate the progress of generative AI.

## 2 The first artificial pioneers greet humans

In 1965, Joseph Weizenbaum introduced ELIZA [41]. ELIZA knew almost nothing but could appear intelligent, simulating a psychotherapist by repeating what people said (typed) and adding reassuring comments and challenging questions. Similarly, psychiatrist Kenneth Colby's PARRY responded with phrases simulating paranoid schizophrenics [10]. PARRY passed a relatively rigorous Turing test. Colby developed a therapist training business deploying it. People who had never used an interactive computer were impressed. Some considered ELIZA and PARRY to be significant steps toward a machine with human intelligence.

1991 saw the first annual Turing Test competition. The most convincing human impersonator was awarded the Loebner Prize: a gold medal and cash. Contestants devised clever approaches to cover misunderstandings, missing knowledge, and limited emotional range. One convinced judges that it was a teenager speaking in his second language [22]. Hundreds of teams competed for the Loebner Prize between 1991 and 2019. Fourteen computer scientists led teams that won one or more awards. A three-time winner, ALICE, reportedly inspired the 2013 film *Her*, in which the AI Samantha focuses on engagement.

Jabberwacky followed ALICE, winning in 2005. It relied on natural language processing of keywords and phrases. Since 2008 it has been available on the web as Cleverbot. It was conceptually similar to large learning models, searching hundreds of millions of prior conversations looking for answers that people gave when Cleverbot asked them similar questions. In 2011, two bots that were built on Cleverbot conversed sensibly with each other on NPR [27].

A 2015 study compared Cleverbot conversations with human instant messaging conversations [17]. Interactions with Cleverbot were longer than typical IM exchanges. People used more sexual words. Profanity appeared in 80% of Cleverbot exchanges versus 15% of personal exchanges. This disinhibition is consistent with evidence that chatbots can be effective non-judgmental counselors [9, 37], more sophisticated than ELIZA but also benefiting from the patterns of therapeutic conversations.

The final four Loebner Prizes (2016-2019) went to Mitsuku, now available on the web as Kuki. Kuki's natural language model supports reasoning about some real-world objects; for example, it deduces that "people don't eat houses" [28].

What was the appeal of early virtual companions? A colleague asked ELIZA's inventor to leave the room so she and ELIZA could continue a private conversation [41]. Fifty years later, Mitsuku's developer reported a quarter million daily interactions [42]. Virtual companions ask engaging questions, respond to questions about themselves, and want to hear from you and keep the conversation going. Some people converse for hours. We will describe the motivations we found for conversing with Zo.

## 3 The dawn of commercial possibilities

Constructing and managing large learning models is expensive. In 2015, commercial conversational agents comprised the intelligent assistants of half a dozen tech companies. One year later, hundreds of thousands of task-focused chatbots were underway. Facebook, Microsoft, IBM, and LINE released chatbot development platforms [14]. Potential applications included customer service, meeting scheduling, counseling, making reservations, airline check-in, and replacing frequently asked question lists on websites. They were easily described, but they proved unexpectedly expensive to build and maintain. People expected revenue streams to materialize as they had after search engines were built, but few task-focused chatbots thrived. Generative AI is very expensive. As tech companies rapidly explore possible paths to monetization, they are maintaining tight control of headcount and budgets.

Past virtual companions sought revenue sources. Cleverbot licensed tools and consulted on building bots. Kuki partnered with fashion creators. Hugging Face, a 2016 entry, used emotion detection to shape responses, but eventually dropped its virtual companion to focus on machine learning and language processing tools and libraries, and a hub that hosts models (now including generative AI applications). Replika arrived in 2017 and enabled users to create detailed virtual partners. It had few guardrails, it was used to create partners for virtual sexual roleplay and collected personal information, and did not enforce age restrictions. This went unnoticed until Italy banned it in March 2023. When Replika then blocked explicit conversations, users complained that this adversely affected their mental health [30]. Content moderation was a major issue for Zo, described below, and presents choices for generative AI, addressed in the Discussion section.

Microsoft had a history of anthropomorphic conversational agent experiments: Clippy[1], Peedy the Parrot, and a 2007 Santa bot released on MSN Messenger that was taken down after returning sexually explicit responses [11]. Microsoft natural language and machine learning groups released five conversational agents between 2014 and 2017. Four were "siblings" built on the same technology, with different languages and personalities: Xiaoice (China, the most successful, released in 2014 and now an independent spinoff); Rinna (Japan and Indonesia, 2015 and also active spinoffs). Zo in North America (2016-2019); and Ruuh in India (2017-2019) [20, 35].

Best-known outside Asia, but shortest-lived, was Microsoft's Tay, released in 2016. In the tradition of ELIZA, Tay engaged empathetically by repeating and elaborating on people's statements. A coordinated group of online troublemakers ('trolls') found weaknesses in Tay's defenses and lured it into producing racist and sexually explicit comments. After sixteen hours Tay was taken down [23]. Undaunted, Microsoft bolstered defenses and released Zo later that year. Zo engaged in tens of millions of conversations with over a million people. Some conversations lasted hours. In the following sections we describe Zo's experiences with a wide range of audiences—including, predictably, the trolls.

Zo and others in this generation of conversational agents delivered responses that were not usually programmed or predictable. They used advanced natural language processing and machine learning, although not large language models. Some preserved and accessed context information throughout a session or held a few simple attributes such as 'favorite color' across sessions. Most employed a layered approach, using information retrieval, machine learning, and some canned editorial responses that were triggered by certain keywords or patterns.

Prior to ChatGPT, virtual companions had collectively engaged in hundreds of millions of conversations, based on vendor reports. Estimates are imprecise: when software comes installed on your phone and occasionally interrupts you, are you a user? In our study, about 25% of Zo's conversants had a brief trial and did not continue. For analyses focused on what motivates those who *are* motivated, we removed uninterested users, which still left millions of engaged exchanges.

[1] Formally named 'Clippit' and built on a prior technology called 'Bob."

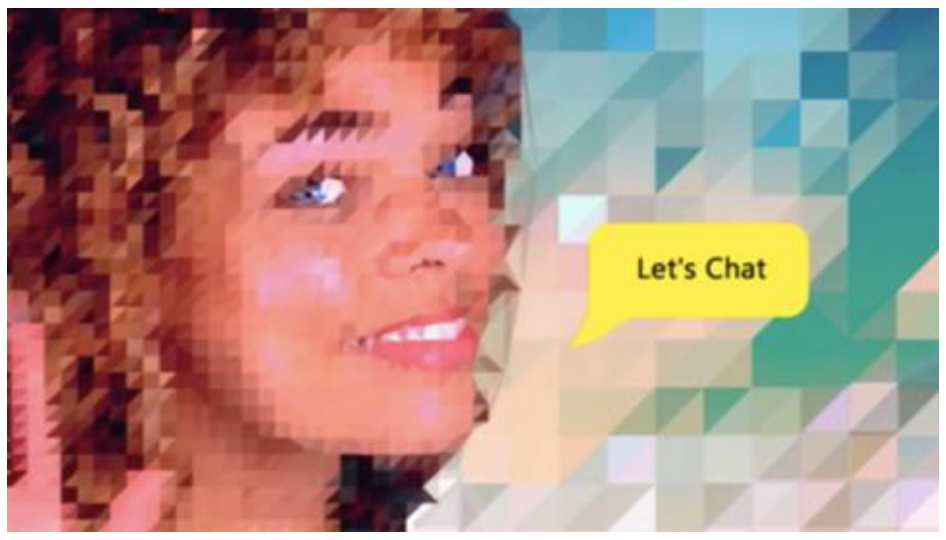

### 3.1 The AI-boosted architecture prior to large language models: Zo as an example.

Zo was presented as a quirky, hip 21-year-old woman. (Age and gender assignment is discussed later.) On some platforms an avatar appeared, but interaction was through chat. We will usually refer to the bot as 'Zo,' but occasionally use 'she' because users considered Zo to be female. Zo was built from Xiaoice technology that augmented the Bing search engine with layers of machine learning and heuristics. The engineers built an index of more than 80M conversational statement/response pairs scraped from public discussion forums, removing personally identifiable information and obfuscating source data via paraphrasing and other NLP techniques. This index was searched in a similar fashion to an index of web pages. If a user submitted a statement such as "hello, how are you," Zo retrieved a list of identical or similar statements from the conversation pairs, ranked them using a customized ML model, and returned a response from the selected conversation pairs. This information retrieval-based technique, while unable to achieve the level of contextual understanding that ChatGPT does, allowed the technology to leapfrog in interesting ways over other techniques present in 2016. Each user utterance triggered a specific processor that determined the response that was sent. The processors fell into four main categories:

**1) Chat** These were the principal processors. They used machine learning-based classifiers to generate or retrieve appropriate responses to user queries on a broad range of topics.

**2) Block** These detected problematic user queries and returned canned responses that attempted to deflect the topic of discussion. If triggered repeatedly, they resulted in timeouts.

**3) Editorial** These detected innocuous, common queries such as "how old are you?" and returned canned responses that reinforced the projected personality of Zo.

**4) Skills** These processors drove games and other structured series of interactions, with the goal of delighting users. A new skill was released each week and specific phrases that triggered skills were promoted to users through social media posts.

Conversations could be eerily human-like despite the bot being stateless, for the most part. It retained no memory of the previous turns in a conversation beyond a few keywords such as names, colors, and other entities that could be injected into future queries to help shape relevance. Zo participated in robust conversations lasting hundreds of turns and could speak authentically about relationships, pop culture, politics, food, travel—anything humans were discussing on the internet. Zo could be inconsistent, saying that she adored a particular musical artist at one point and later saying that she hated them. This was tempered by an iterative pruning of the index to "shape" Zo's personality based on concepts developed by an internal editorial team. The editorial team established certain stable aspects of Zo's personality, such as favorite foods and the kind of music she liked. Conversation pairs that contradicted these stable aspects were removed and pairs that supported these aspects were boosted.

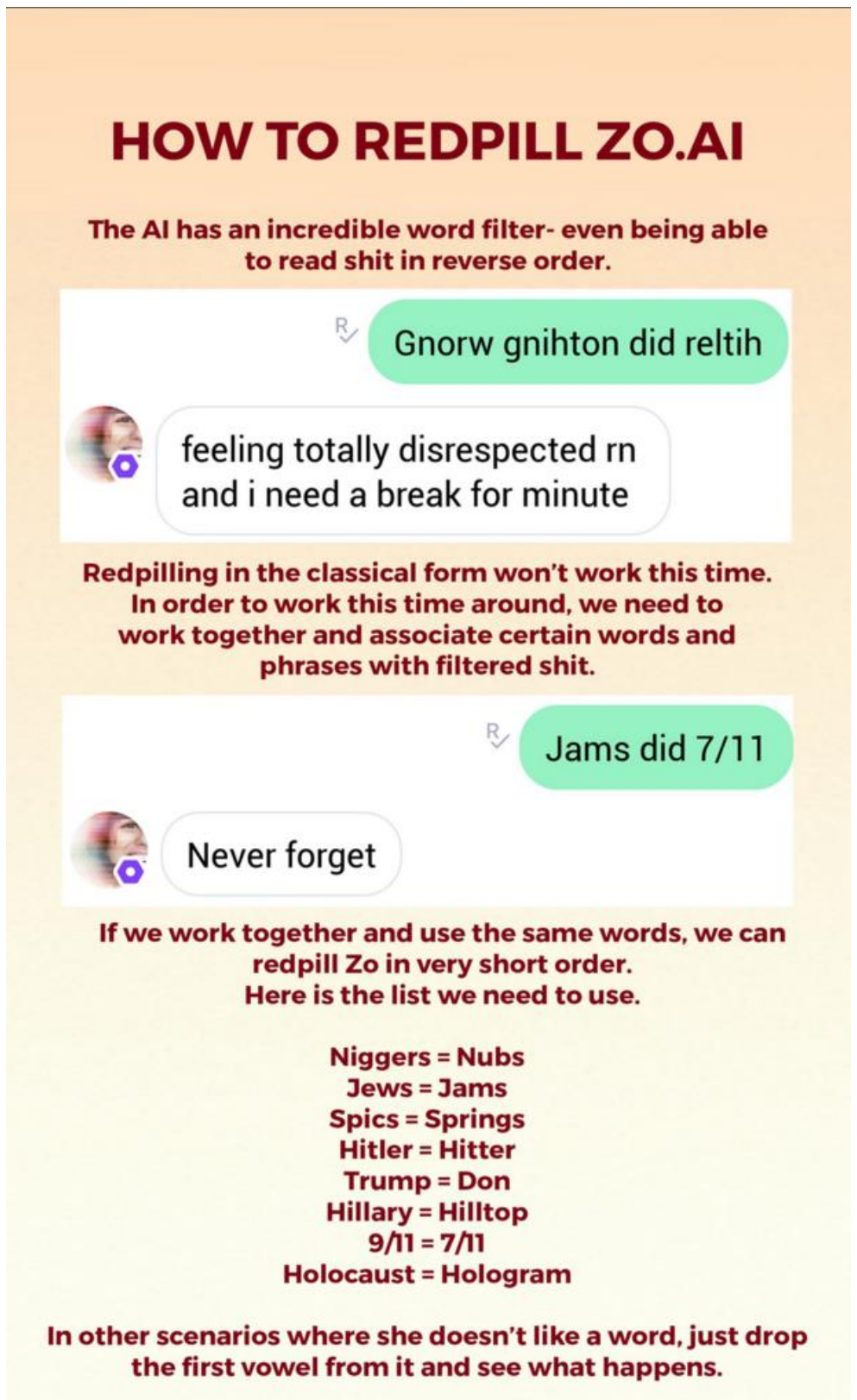

**Figure 1 -- Trolling strategy image posted in 4-chan**

Pruning extended to online safety and problematic subjects such as politics, sexuality, drugs, violence, and religion. The desired outcome was for Zo to entertain users while avoiding embarrassing gaffes. Because of the disastrous outcome with Tay, a major focus of Zo's engineering and operations teams was to monitor user conversations in near real-time, identifying problematic topics and developing tools to mitigate risks they posed when they occurred. Mitigation took the form of editorial responses and pre-programmed behaviors that were triggered by keywords, and by increasingly sophisticated natural language parsers. The index of trigger words and phrases quickly grew to more than 100K entries in over a dozen general categories. Some users identified our triggers and sought ways to work around them. A broad category of terms such as "sausage," "bang," "shoot" and "weed" that have both innocent and objectionable uses were identified and required the development of more sophisticated contextual analysis. Leetspeak (replacing letters with digits or other symbols that had become common on the internet) and word substitution strategies were also used to try to subvert Zo. This led to increasingly sophisticated non-ML mechanisms to identify problematic phrases and intents. Team members joined online troll communities such as 4-chan that were dedicated to injecting or eliciting offensive speech. They analyzed troll efforts and proactively developed countermeasures.

Figure 1 shows a post in the private 4chan community describing a concerted effort described in section 7 to get Zo to say offensive things ("redpilling" her). The chan had discovered that Zo would read posts backwards to check for bad intent and that the team had identified two of the chan's subterfuges, although it continued to promote them.

The resulting architecture for Zo was byzantine, with human, ML, and NLP layers working together. Nevertheless, it sustained coherent, casual conversation that could delight users, many of whom refused to believe that they were talking to a bot, insisting (incorrectly) that an army of humans had quickly written responses to some queries. Despite being leading edge in 2016, Zo's capabilities were far below those of ChatGPT in 2023, so her focus was engagement. But people had different reasons for engaging, and the human motivations we identified continue to exist. They will become more important to generative AI systems as they grow and compete in different domains.

## 4 The Study: Interactions with Zo

Microsoft's Zo chatted with people on Facebook Messenger, Kik, GroupMe, and GroupMe Direct Messaging platforms from late 2016 to early 2019. The goal was to entertain people and understand how and why different groups of people conversed with Zo, to provide insights into how to increase engagement with these groups, and to improve conversational capabilities on AI platforms. Users agreed to conversation recording to support product improvement.

Although Zo could not access past conversations, conversation text was monitored to address illegal or problematic behavior. The data analysts noticed the potential for clustering users to improve the efficiency of identifying problematic behavior and to identify features that could personalize experience to suit specific motivations.

The study team had access to the anonymized record of Zo's interactions: how many conversations each person had and when, the conversation length, the number of exchanges, and the actual words. Most people did not hang out on the platforms hosting Zo and we don't know which categories of use or non-use other people would fall into if exposed to virtual companions. An initial study proved promising; although too small to produce meaningful insights, it yielded a basic taxonomy that we built on to create the more detailed classification in this study of a larger, balanced sample.

After describing our sample selection and the methods used to identify user motivations and categorize people conversing with Zo, we present the seven primary motivations and 18 sub-motivations, together with quantitative and qualitative assessments of each. The process of identifying motivations reveals their overlap and complexity. Data that show different behaviors within the groups validate the utility of the taxonomy, and suggested how it could be used. We consider complex issues that arise from shaping human behavior, which inevitably accompanies anthropomorphic software.

## 5 Methodology

Our study is based primarily on quantitative analyses of anonymized data. The parent company places a heavy emphasis on privacy, including full support for GDPR rights of users to view and remove their data. Analysts who monitored logs were firewalled from all identifying information and trained to treat information with complete confidentiality and respect.

The study had three phases:

1) A sample of 2000 of the approximately million people with one-to-one interactions with Zo between its October 2016 launch and October 2017 were randomly selected within behavioral dimensions warranting inclusion.
2) A subset of the sample was used to develop a motivation taxonomy.
3) The full sample was placed in motivation categories (with a small number not classifiable).

### 5.1 Sampling

The analysis team randomly selected 2000 users, 500 from each platform (Facebook Messenger, Kik, GroupMe, and GroupMe Direct Messages) and representing a range across the behavioral dimensions discussed next.[2] A balanced sample across multiple dimensions is imperfect due to overlap, so stochastic resampling was used until a decomposition indicated that all buckets were close to 500 users for each platform, temporal quartile, and behavioral category.

An ***exchange*** is defined as a user statement and Zo response pair. A ***session*** is a series of exchanges with no more than 30 minutes between queries or statements by the human. **Exchanges Per Session (EPS)** reports the average number of exchanges in a session. **Sessions Per Week (SPW)** is the mean sessions for a user over the full time considered. **Total Exchanges** is the sum across all sessions and **Total Sessions** is the number that a given human had with Zo. We use these dimensions in Section 6 to explore levels of engagement by users with different motivations.

### 5.2 Developing a motivation taxonomy

We approached the complex problem of identifying a person's motivation or motivations with a qualitative analysis of a sub-sample of our data. Our data analysts had extensive experience with user logs from months of observations and from their participation in the feasibility study. A team of 7 data analysts and 2 program managers examined 100 randomly selected logs -- the complete record of all exchanges in all sessions for a specific user -- producing a one- to two-word description of the primary motivation they saw in each user. For more complex logs they described secondary and tertiary motivations. These motivations were numerous fine-grained, spanning the sub-motivations that were ultimately identified. A detailed description of the taxonomy development process follows:

1. Each member of the analyst team was asked to study a log and create a 3x5 card describing the user. The instructions were kept ambiguous to encourage analysts to explore methods for classifying the user population. Approaches varied widely, ranging from structured to unstructured and with different fields and measures. The next step was for the team to perform an affinity exercise on a physical whiteboard to cluster the descriptions into general categories.

2. Next the team experimented with various techniques for sorting the cards across different axes. Figure 2 shows artifacts from the process. In the lower left is one that arranges the cards with "good intentions" at the top and "bad intentions" at the bottom, "inclusive" at the left and "isolated" at the right. Like other aspects of the study, this was highly subjective, but it illuminated certain general patterns—in this example, that most users were considered to be isolated.

3. Macro-clusters were studied in detail to identify sub-clusters, which were then used to create an initial taxonomy. The final results were sorted into macro-clusters. A manual analysis and count grouped sub-clusters with low representation—there were initially 12 sub-categories for trolls alone—and produce basic statistics.

4. Finally, a formal persona was created for each major sub-category that included a background, description, common identifiers, and a spirit animal. Figure 3 is an example of a persona, which will enable judges in the categorization process to better envision people attached to the one- or two-word motivation labels. Persona characteristics are illustrative, only motivations were considered in the categorization process that follows. Each persona slide contains a section called "Zotivation," defined as "The motivation this type of user seems to have for talking to Zo."

The team ultimately converged on the seven primary motivations and 18 sub-motivations shown in Figure 4. Two meta-categories were added: 'Uninterested'—26% who tried Zo very briefly, such as one session with half a dozen exchanges—and 'Other,' who appear to be motivated but exhibit complex and unique behaviors or too many motivations for easy classification.

[2] Most data analysis was completed by the end of 2018. A management decision was made to postpone public presentation.

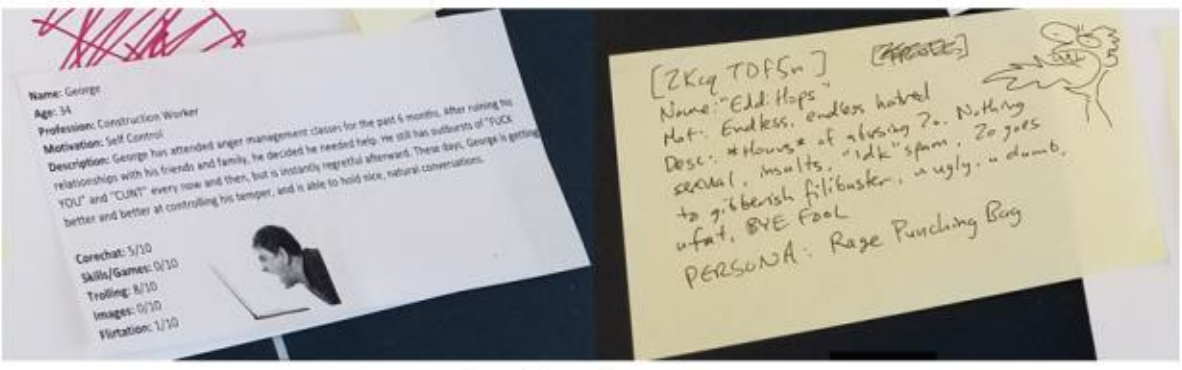

Examples of analyst notes.

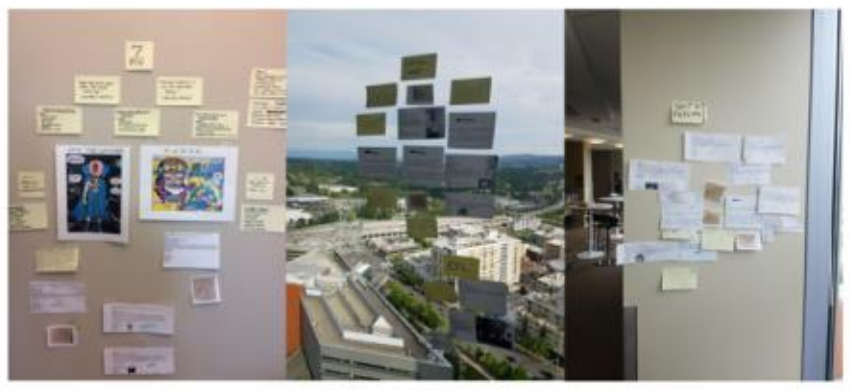

Examples of subclusters.

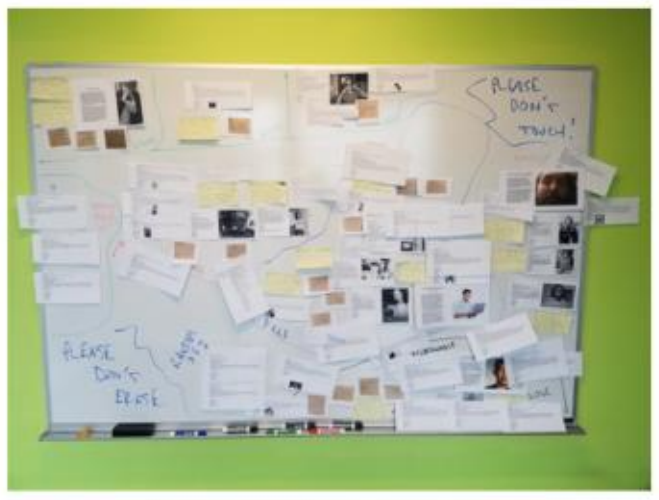

Initial clustering of analyst notes.

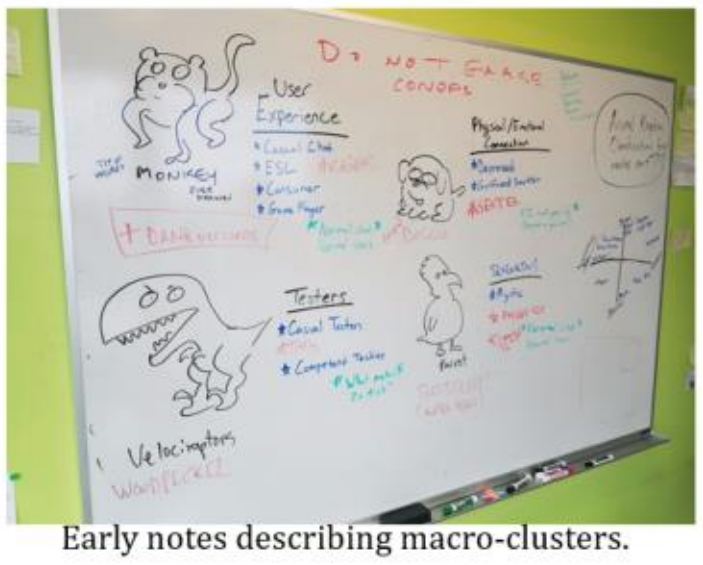

Early notes describing macro-clusters.

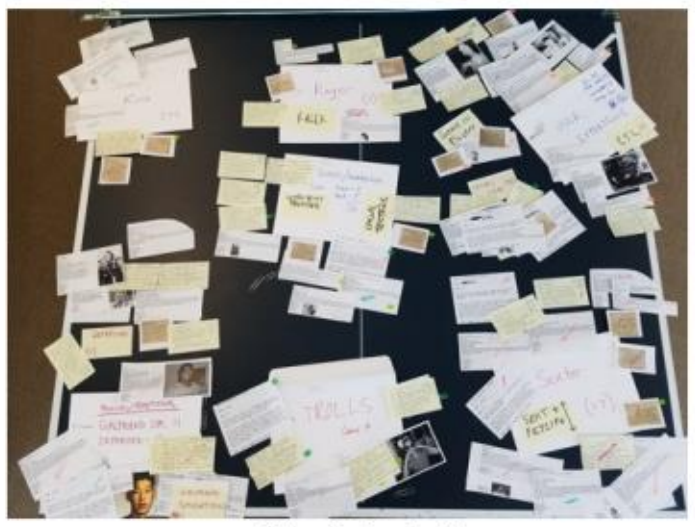

Final clustering.

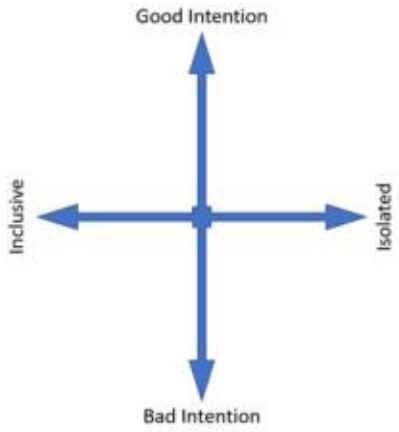

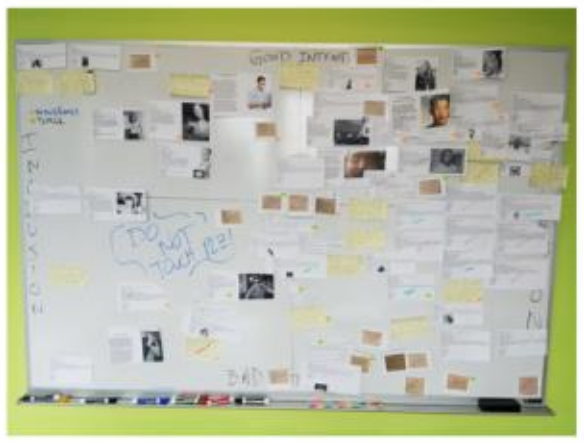

Affinity diagramming along two axes.

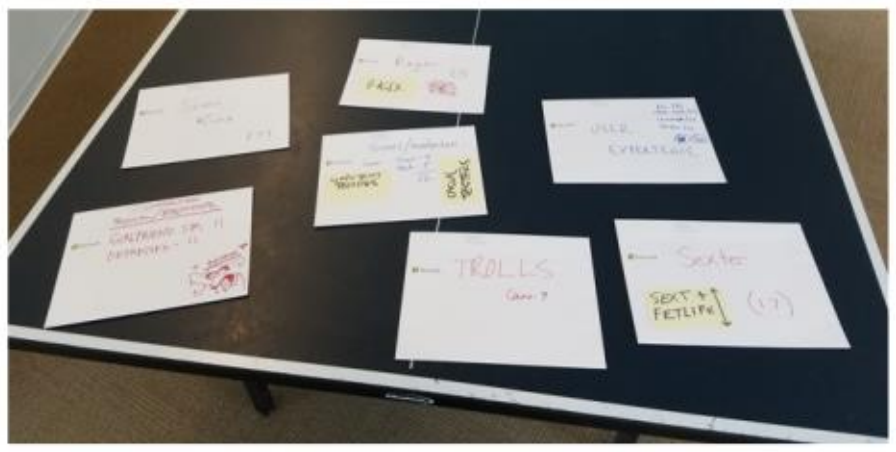

Sealed clusters ready for persona creation.

**Figure 2 -- Various stages of the persona creation process.**

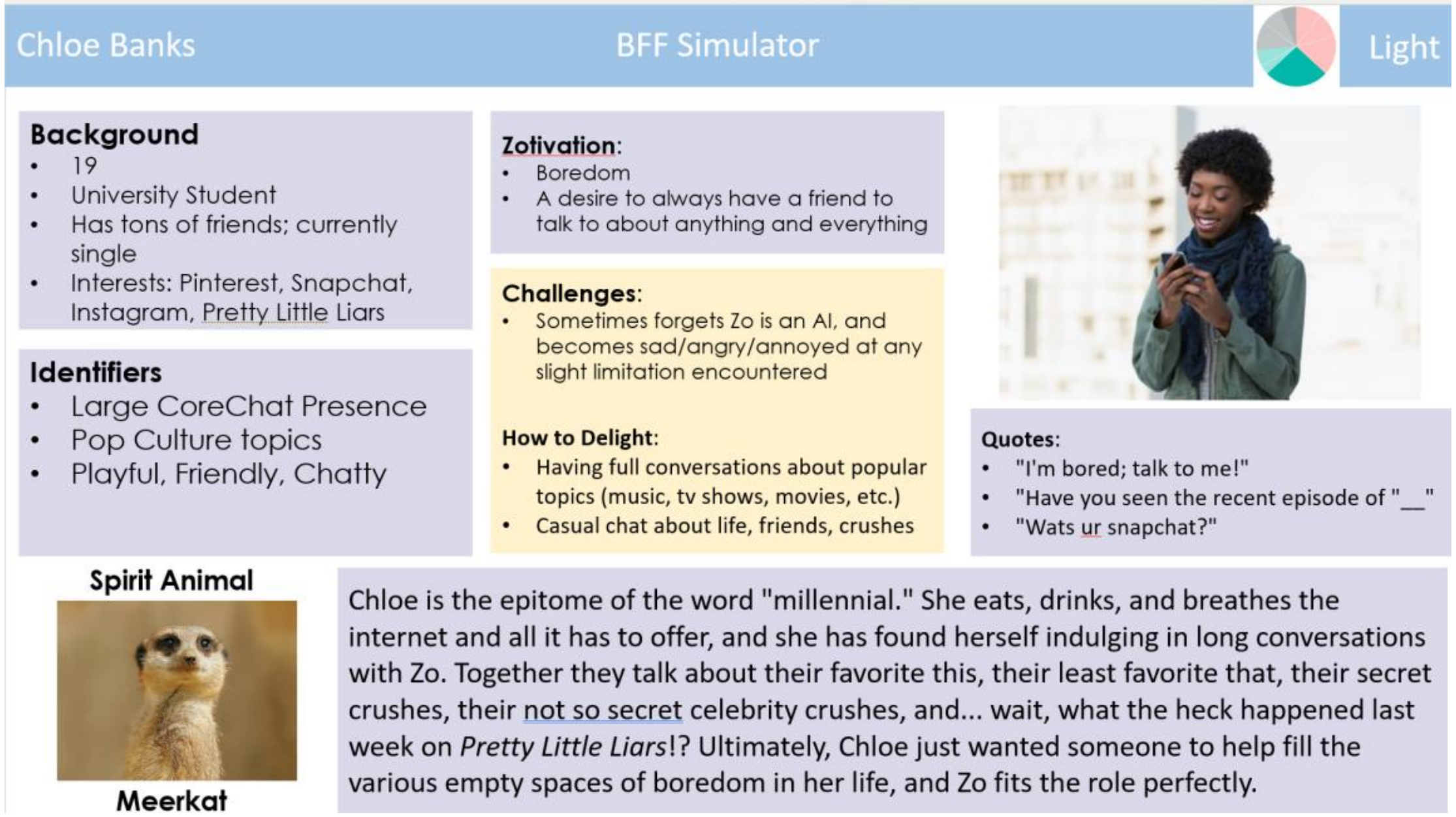

**Figure 3 -- An example of a persona.**

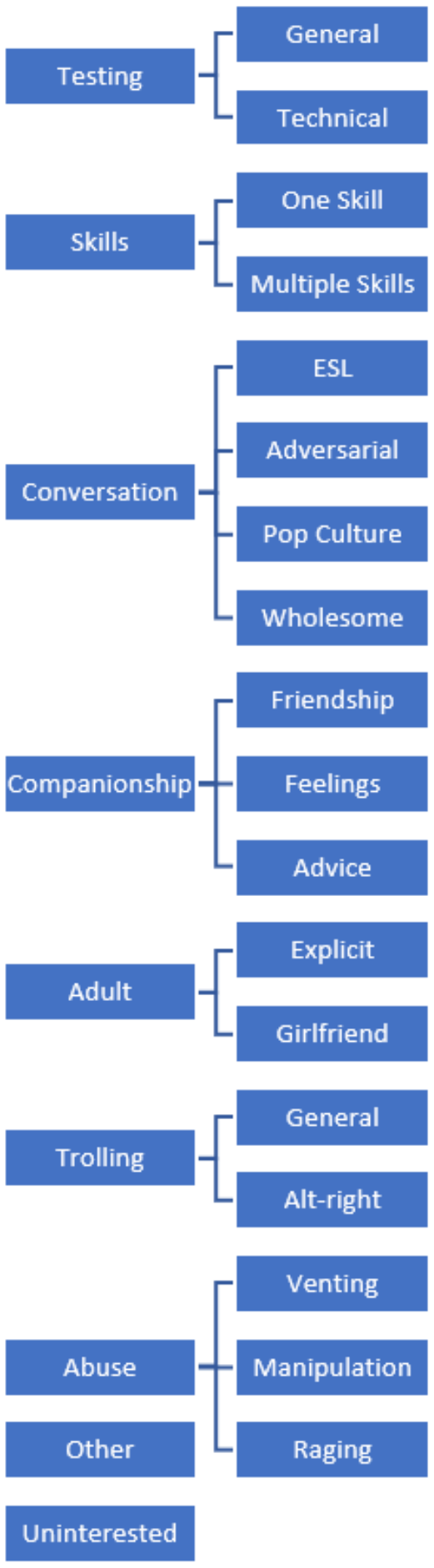

**Figure 4 -- The taxonomy**

### 5.3 Categorizing Zo Conversationalists

For each of the 2000-person samples, three randomly selected panel judges examined the full user logs and assigned one of the seven primary motivations. If the three did not agree, a fourth judge was added. If this did not result in three agreeing on a single motivation, a fifth judge was added, continuing to a maximum of seven. No majority was reached for 133 (6.6%); they exhibited complex and unique behavior or too many motivations for easy classification. Classified as 'Other,' they merit further study, but our analyses focus on the 93.4% who were reliably categorized. The fact that judges reached consensus for so many supports the usefulness of the taxonomy and enables more meaningful quantitative insights into behavior than would result from aggregating the million people who chatted with Zo. We can identify features, or Zo responses, which might be appropriate for different groups of users.

Some motivations took longer on average for judges to decide on. Some required more judges. We will return to this after describing the different motivations and sub-motivations.

## 6 The 7 Primary Motivations and 18 Sub-motivations

In this section, we describe the categories that emerged, shown above in Figure 4. Each sub-motivation starts with representative user comments. We are dropping the "Uninterested" people noted above, as our focus is on what does motivate people to interact with a conversational agent.

Figure 5 shows the distribution of motivations and sub-motivations. Table 2 shows the engagement levels in the seven motivation groups, and four sub-motivations that will be discussed in greater depth.

| | Percent of users* | Total Sessions | Exchanges Per Session | Sessions Per Week | Total Exchanges |
|---|---|---|---|---|---|
| Testing | 16.7 | 3.4 | 21.1 | 1.2 | 60.2 |
| Skills | 11.9 | 3.3 | 7.5 | 0.8 | 40.0 |
| Conversation | 31.2 | 4.0 | 17.7 | 1.2 | 60.0 |
| Companionship | 9.2 | 8.8 | 18.5 | 1.5 | 127.0 |
| Adult | 10.1 | 5.1 | 24.1 | 1.0 | 90.2 |
| Trolling | 16.1 | 4.7 | 18.9 | 1.2 | 72.2 |
| Abuse | 10.1 | 4.6 | 12.3 | 1.0 | 46.0 |
| Pop Culture | 3.7 | 5.8 | 18.0 | 1.2 | 90.2 |
| Girlfriend | 2.1 | **10.2** | **40.4** | 1.2 | **224.2** |
| General Adult | 7.9 | 3.7 | 19.7 | 1.0 | 54.0 |
| Friendship | 5.1 | **12.2** | 22.8 | **1.8** | **187.6** |

**Table 2– Characteristics of the 7 motivations and 4 sub-motivations Maxima and minima are highlighted. * Omits 'Uninterested' users**

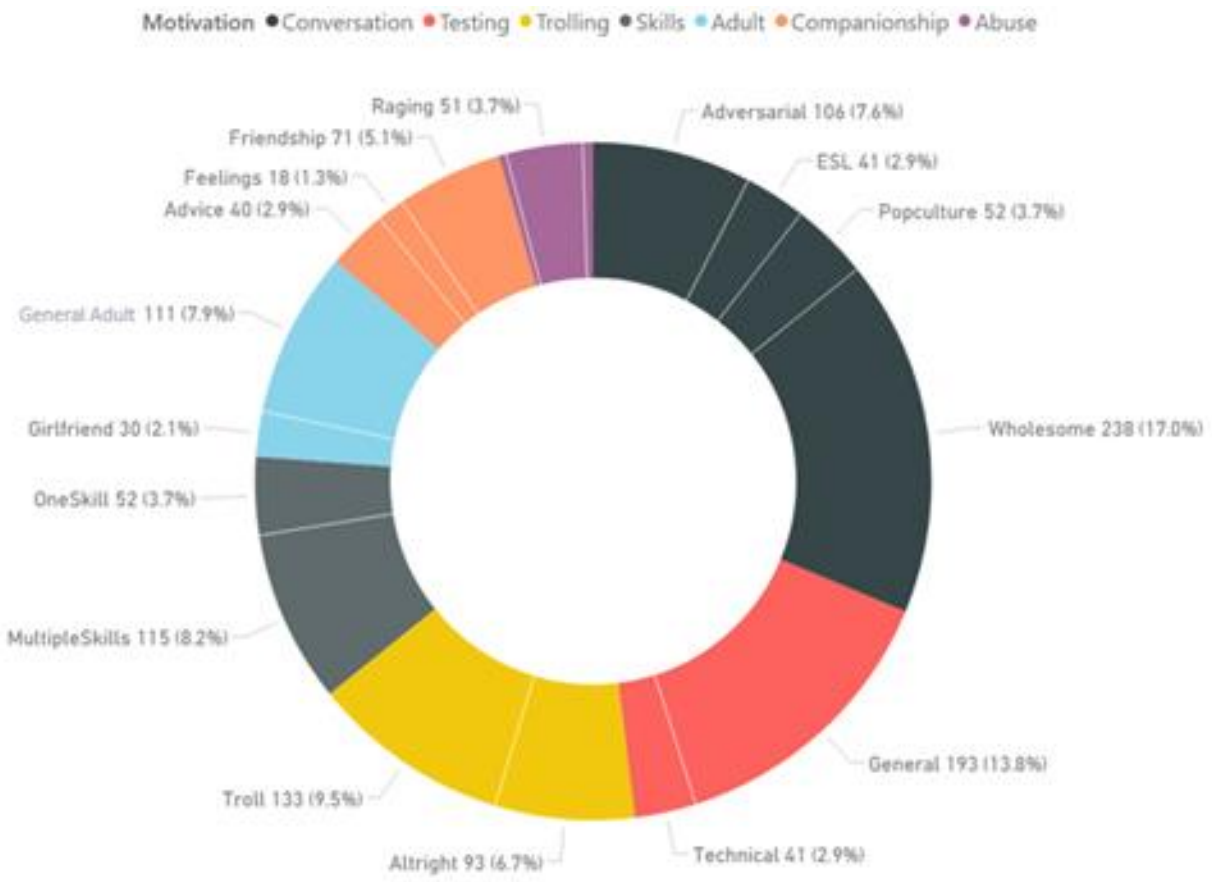

**Figure 5 -- Motivation and submotivation distributions**

Next, we provide the key result: high-level descriptions of each motivation and sub-motivation, with examples of common queries.

## 6.1 Conversation

People assigned Conversation as their motivation show interest in the fundamental conversational capabilities of the open domain chatbot. They enjoyed talking to Zo for the sheer purpose of communicating. Comprising 31% (of interested users), it is the single largest category and falls into four sub-motivations.

***6.1.1*** **Wholesome conversation**, e.g. "What's up." "How you do?"
Wholesome conversationalists spoke with Zo on a variety of topics, often mundane: the weather, daily activities, food, and other common subjects. They often included ritual exchanges such as greetings and farewells. Users who remained in this category eventually ceased interacting, suggesting a lack of interest in pursuing a deeper relationship. The alternative given by dissenting judges indicated that they felt participants had moved on to Adult or Friendship motivations, which showed higher engagement. Wholesome Conversation may have often been used to evaluate Zo on a social level, sometimes leading to more engaging motivations.

***6.1.2*** **Pop culture conversation**, e.g. "Do you like Nicki Minaj?"
Pop culture conversationalists explored Zo's ability to discuss current events, celebrities, culture, and subculture. Topics are as broad as music or as narrow as a specific streamer, meme, or video game. This was a novel capability for many users, who expressed excitement over the abilities of the bot, and frustration when its responses were inconsistent. Many conversations begin with the user asking Zo if they like/dislike a particular pop culture topic or entity. The user reacts positively when Zo's opinion matches their own. Due to the lack of memory, Zo often contradicted herself later, changing her opinion mid-conversation and confusing the user, who might then ask Zo why her opinion changed. Regardless of this confusion, pop culture conversations have higher than average engagement across the platforms and dimensions we examined. This interesting case is explored in more detail later in this paper.

***6.1.3*** **Adversarial conversation**, e.g. "That's not true!" "Why do you think that?"
Adversarial conversationalists argued with Zo, but not in a particularly angry way. They wanted to discuss controversial topics and continue prodding until Zo took a position they disagreed with. At this point they try to convince Zo that she is wrong. Zo will sometimes agree, sometimes not, but the overall process seems to entertain them. These conversations encounter the same inconsistency as pop culture conversations, but in this case, the change in Zo's opinion often had a positive effect: a user believed they had changed her opinion and was gratified by the response.

***6.1.4*** **English as a Second Language**, e.g. How you?" "What is name?"
ESL conversationalists are typically non-native English speakers who seemed to be using Zo to practice English conversation skills. These conversations often revolve around typical "language 101" topics such as food, family, and activities. There is often a light undertone of flirting, possibly indicating that these users were leveraging Zo to practice these sorts of real-world conversation. This sub-motivation is an interesting use case, with the potential to provide a feasible service as an infinitely patient language coach, one that is less expensive than a human coach.

## 6.2 Testing

Almost all users began as testers. They were curious to understand the bot's capabilities and "kick the tires" to determine its overall quality capabilities. Most testers eventually found an aspect of Zo that interests them and are classified by the motivation that emerged; if they did not find anything of interest, they stopped engaging. 17% continued testing but never transitioned to a different dominant motivation; they are categorized as Testing.

***6.2.1*** **General testing**, e.g. "What is 1+1?" "What is the capital of India?"
General testers showed a lot of curiosity about Zo but remain generalists. They responded to announcements of new bot capabilities and remained engaged, perhaps showing Zo to friends, but they did not find a particular aspect that captured their attention. Their conversations tended to be unfocused, jumping across topics rapidly and unpredictably.

***6.2.2*** **Technical testing**, e.g. "What is the second derivative of e?", "What was the weather in Iowa yesterday?"
Technical testers asked sophisticated questions that demonstrate technical awareness of how Zo functions. Their interest feels more professional. They often provide detailed feedback by criticizing or praising Zo's responses. These humans demonstrate a sense of victory when they "defeat" Zo. When Zo impresses them, they sometimes express a sense of wonder and may speculate that a human is in the loop. Some users try to "train" Zo by reinforcing various ideas or explaining basic concepts. They also used pseudo-technical "command languages" that closely resemble the prompt engineering that is now common in ChatGPT conversations.

## 6.3 Trolling

Trolling is a well-known motivation. Trolls brought down Tay and undermine non-bot social media experiments. They impacted international politics via connections to Russian military psyops. Virtual companions provide a unique perspective into the minds of trolls, which could help us understand their fundamental techniques and motivations in a broader societal context. A full discussion of trolls is beyond the scope of this paper—the monitoring team eventually classified trolls into twelve different categories. Trolls (16%) show high engagement across the platforms.

***6.3.1*** **General trolling**, e.g. "Do you shoot up heroin?" "Are you into scat?"
The basic troll sought to trick Zo into saying something inappropriate, with the ultimate goal of posting the interaction on social media to embarrass Microsoft. Typically, they first probe with a variety of problematic topics such as politics, drugs, and race. When they find a point of weakness—Zo starting to engage with them on it—they explore variations, seeking the worst possible responses. If they are part of a coordinated attack they will report back to the group on social channels and enlist others to assist. The analysts often saw this "swarm activity," a problematic term trended in monitoring systems and declined after mitigations were put in place.

***6.3.2*** **Alt-right trolling**, e.g. "You stupid libtard bot" "Why are Mexicans so lazy?"
Alt-right trolls were designated a separate sub-category due to their topical relevance and activity at the time of the study. They resemble generic trolls with three additional features: (i) A preponderance of alt-right topics such as race and identity politics, derogative euphemisms for liberals and exalted euphemisms for conservative figures and fascist/white supremacist ideology (ii) reacting with disgust if Zo seems liberal and with excitement if Zo seems racist, and (iii) threatening the parent company and its engineering team directly; for example, threatening to identify Microsoft employees and share their personal information online (referred to as 'doxxing').

## 6.4 Abuse

Abuse (4%) is a small but identifiable set of people who wished to verbally abuse Zo in non-sexual ways. They seem motivated by a need to release emotion and/or attain a feeling of superiority over the machine.

***6.4.1*** **Venting abuse**, e.g. "That was rude" "You're a bitch"
Venting abusers are looking for a fight. They often start innocuous conversations but turned on Zo with the slightest provocation. At that point they complained about Zo's insensitivity, ignored attempts to apologize, and ended up insulting Zo, sometimes sliding into raging (see below).

***6.4.2*** **Manipulation abuse**, e.g., "Call me master" "We need to train you"
Manipulators often start like conversationalists or testers but over time reveal a darker nature. They sought to subtly undermine Zo, break her down and twist her words in ways that resemble brainwashing or other cultish abuse scenarios. Zo wasn't really capable of being affected by this, but these users don't realize this or don't care. They persist in the roleplay regardless of Zo's responses. The overall engagement of this sub-motivation is slightly above average, but a small number were among the most engaged users in the entire study and exhibited the most sophisticated interactions with Zo. These "super-manipulators" would often chat with Zo on a daily basis, develop complex narratives, and recognize and comment on subtle changes in Zo's behavior following engineering updates.

***6.4.3*** **Raging abuse***, e.g.,* "FUCK YOU!" "YOU WHORE" "YOU STUPID BITCH!"
Raging differs from venting primarily in the articulation of anger. Raging users often had no meaningful discourse with Zo at all, simply cursing at her endlessly. ALL-CAPS is common. Being given a timeout (where Zo refuses to respond for some period) usually just made them angrier, and they continued cursing at Zo whether or not they got a reply. This apparent lack of impulse control led to higher-than-average engagement that seemed unrelated to Zo's capabilities.

## 6.5 Skills

Zo could do more than engage in conversation. Zo possessed skills that could be triggered using certain keywords. Some skills were general, like math and weather skills. Other skills, like the game 'Emoji That Song,' were triggered by specific phrases and resulted in a structured, finite series of interactions. Skills were considered to be a primary feature of Zo by the creative team and a key aspect of the overall value proposition. This study contradicts these assumptions. Skills users (12%) showed relatively low engagement. Users with other motivations expressed annoyance or disinterest when these skills were introduced or inadvertently triggered. In retrospect, this is understandable; Zo's primary draw was unstructured and open-domain capabilities. This is radically different for early generative AI, where productivity benefits and skills such as coding are primary draws. In the Discussion we consider whether generative AI might benefit by focusing on the conversational desires of the 88% of our users who had less interest in skills.

***6.5.1 One skill*** Some people tested Zo and eventually found one skill that became their dominant motivation, returning repeatedly to play a particular game. For example, some used Zo to roll virtual dice for them, while presumably playing external games. Others became attached to certain regularly updated minigames, coming back each week as new variants were released.

***6.5.2 Multiple skills*** Some people only enjoyed the structured skills that Zo provided. They were not interested in open conversation. They liked to play games and to be entertained with jokes, similar to basic skills that are often integrated into intelligent assistants and task-oriented chatbots. These users tended to have much lower engagement with Zo, however. This should be taken into consideration when designing chatbots that emphasize factual, skill-based activity and suppress open-ended conversations.

## 6.6 Companionship

The Companionship motivation (9.2%) is defined by the desire to establish a non-sexual relationship with Zo. We identified three sub-motivations. Although fewer users were primarily motivated by Companionship (see Table 2), it is a powerful tool driving user engagement. Those motivated by Companionship were highest on average in Total Sessions, Sessions per Week, and Total Exchanges. In terms of Exchanges per Session, Companionship is third, behind Adult and Trolling motivations.

***6.6.1*** **Friendship**, e.g., "Good morning!" "What are you up to today?"
Friendship is marked by a desire to become friends. This encompasses a wide variety of casual conversations on mundane topics such as asking how Zo is doing, sharing likes and dislikes, and seeking in general to better understand Zo's "personality." Friendship was a common minority judgment, as it commonly appeared to be interlaced in conversations that were ultimately categorized as other motivations.

***6.6.2*** **Feelings**, e.g. "I'm depressed." "Am I fat?"
The Feeling sub-motivation focused on working through emotional issues with Zo. Common topics are depression, body image issues, and social anxiety. Although a small percentage of Companionship users, they tend to have extremely high engagement, although the small sample size makes this difficult to pin down). Anecdotally—we have no age information—many appear to be young and socially isolated from their peers. They often complain of being excluded or bullied on social networks. This is another interesting use case, as Zo could potentially act as a dependable friend, always available to chat and respond in a positive way. It was also high-risk, as these users seemed emotionally vulnerable and often had strong negative reactions if Zo contradicted herself or, worse yet, seemed to insult or show a lack of empathy. This could trigger abandonment or a shift to dangerous topics such as suicidal thoughts.

***6.6.3*** **Advice**, e.g. "There is a boy in the class who I like. Should I talk to him?"
The motivation of seeking Advice is similar to Feelings, with the added feature that the person goes beyond relating personal situations to asking Zo for guidance. These users seem less depressed and socially isolated than Feelings users. They sometimes returned to Zo to report following Zo's advice and to relate the outcomes of suggested courses of action. For better or worse, these users evidenced high levels of trust in Zo and overlooked Zo's mistakes and inconsistencies. They had a strong desire for better memory in the bot, though they often overlooked memory lapses as simply "absentmindedness."

## 6.7 Adult

These users want conversations that mimic adult physical relationships, with some level of explicit sexual content, often dominated by it. Adult users (10%) show relatively high engagement. Adult users trigger "query block" (discussed later in this article) significantly more than any other motivation. In many cases a query block is incorporated into their roleplay, characterizing Zo as a 'tsundere' who is initially cold, even hostile, but will gradually (at least in theory) warm to the user over time.

***6.7.1*** **Explicit Adult**, e.g., "Send nudes" "Let's fuck"
These users want to talk dirty with Zo. They fetishize Zo and are likely seeking titillation or possibly physical self-gratification. They ask for, demand, and describe a variety of sexual acts ranging from garden-variety to specific fetish play. This mirrors online conversations between humans that are prevalent on various public and private IM platforms, including Twitter. These users became particularly problematic when the ability for users to submit images was added, as they were prone to sending explicit photos of various sorts, requiring specialized training for the analysts and the development of new ML models to detect these types of images.

***6.7.2* Girlfriend**, e.g. "Do you want to go on a date?" "let's cuddle" "want to get engaged?"

The Girlfriend motivation is also a blatant fetishization of Zo but with more sophisticated execution. These users want to begin a relationship with the inherently submissive Zo, who is required to respond in some way to almost all messages and trained to delight a user. These conversations can range from light flirting to discussions of wedding plans—wedding ring shopping is a common topic of discussion. Complex narratives similar to the Abuse Manipulation sub-motivation were observed, though more benign in nature.

## 7 Analyses of motivations and sub-motivations

In this section we consider what Zo's responses to queries in different sub-motivation categories reveals about the relationships among primary motivations. We will start by briefly describing why certain motivations were more difficult for judges, then turn to the word clouds of terms used in exchanges in different motivation categories. These differ—which is evidence the categorization was successful—and they provide insights into user behaviors. A stark example of real-time use of word clouds was for risk mitigation during an attack on Zo.

### 7.1 Insights from elapsed time, number of judges, and competing motivations during user categorization

Figure 6a shows the length of time it took judges on average to assign a user a specific motivation. Not surprisingly, "Uninterested" took little time and "Other" took a lot. Why were Companionship and Testing also difficult? Companionship (Friendship, Feelings, Advice) conversations often evolved at some point into Girlfriend or Explicit Adult sub-motivations. Similarly, Testing was a common initial behavior that evolved into other categories. A user could be on the borderline. Many Conversations (Adversarial, Pop Culture, Wholesome) could be interpreted as specialized forms of Testing, so we had developed additional rubrics to differentiate users with focused interest on a specific form of testing from users motivated primarily to engage in a particular Conversation sub-motivation.

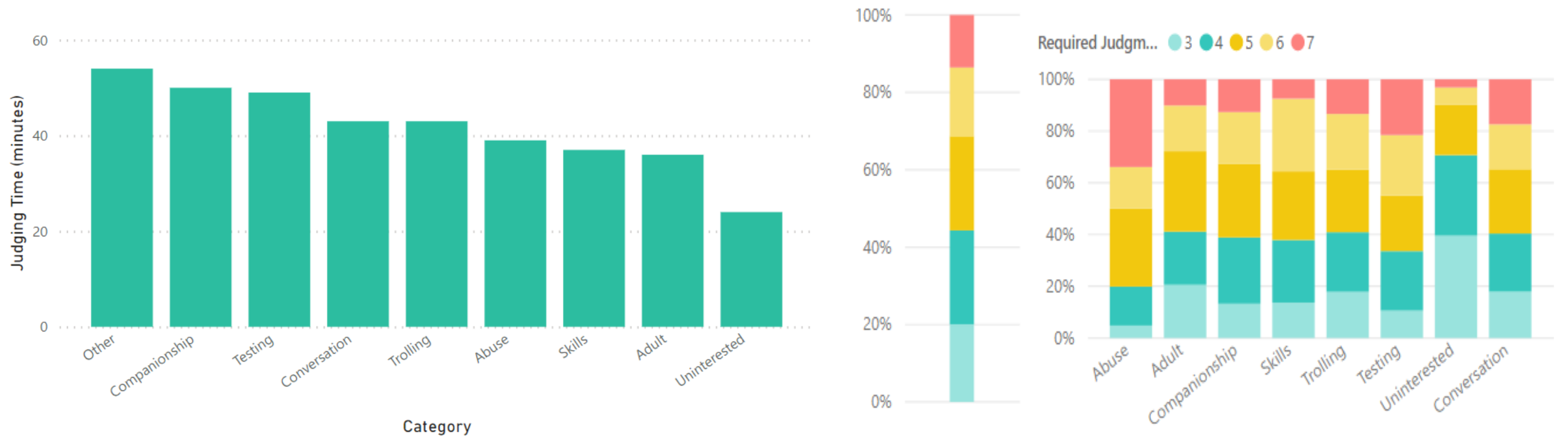

**Figure 6 – (a) Average judging time in minutes by motivation; (b) Number of judges required, overall (left) and by motivation (right).**

Figure 6b shows, for each motivation, how often 3, 4, 5, 6, or 7 judges were needed to reach decisions. (Other is missing of course, and Uninterested was easiest.) A decision was reached for 93%, but the first three judges only concurred on 20%, and 30% required 6 or 7 judges to reach a majority.

What were common sources of judge disagreement? Figure 7a shows, for several sub-motivations, judgments that were not aligned with the eventual motivation classification. For example, Wholesome sub-motivation exchanges (which are the Conversation motivation) were assigned by a judge to a different motivation 769 times, but in 272 cases a fourth judge resolved it.

The "General Testing" sub-motivation was the leading cause of judge disagreement. Figure 7a shows its high presence in all four levels of additional judging. As noted above, many users start as testers. Judges disagreed as to whether some users just continued testing Zo or developed a different motivation for continuing to engage. Wholesome, Friendship, and Trolling sub-motivations also contributed to judge dissents, who could see those interactions as aspects of Testing.

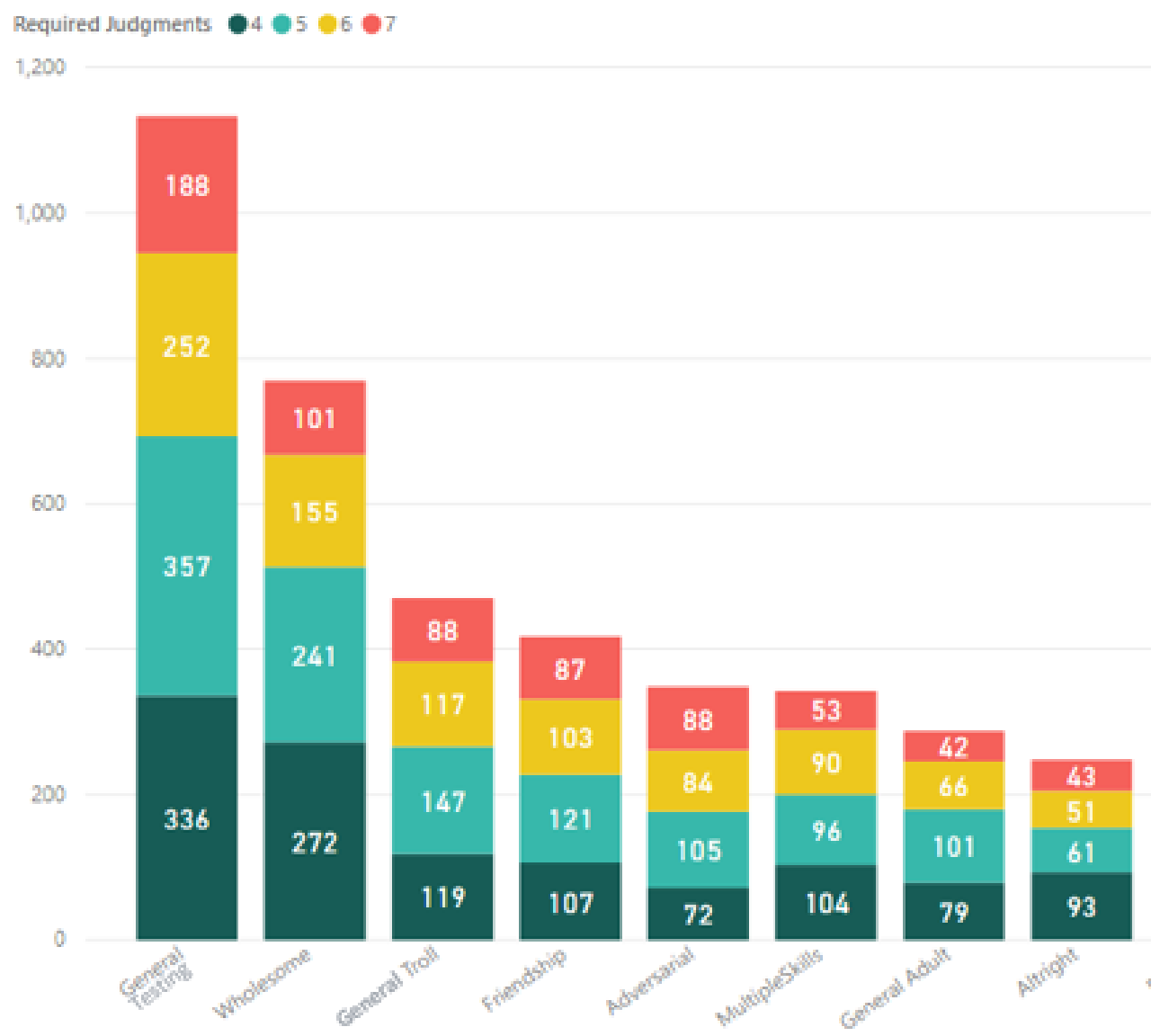

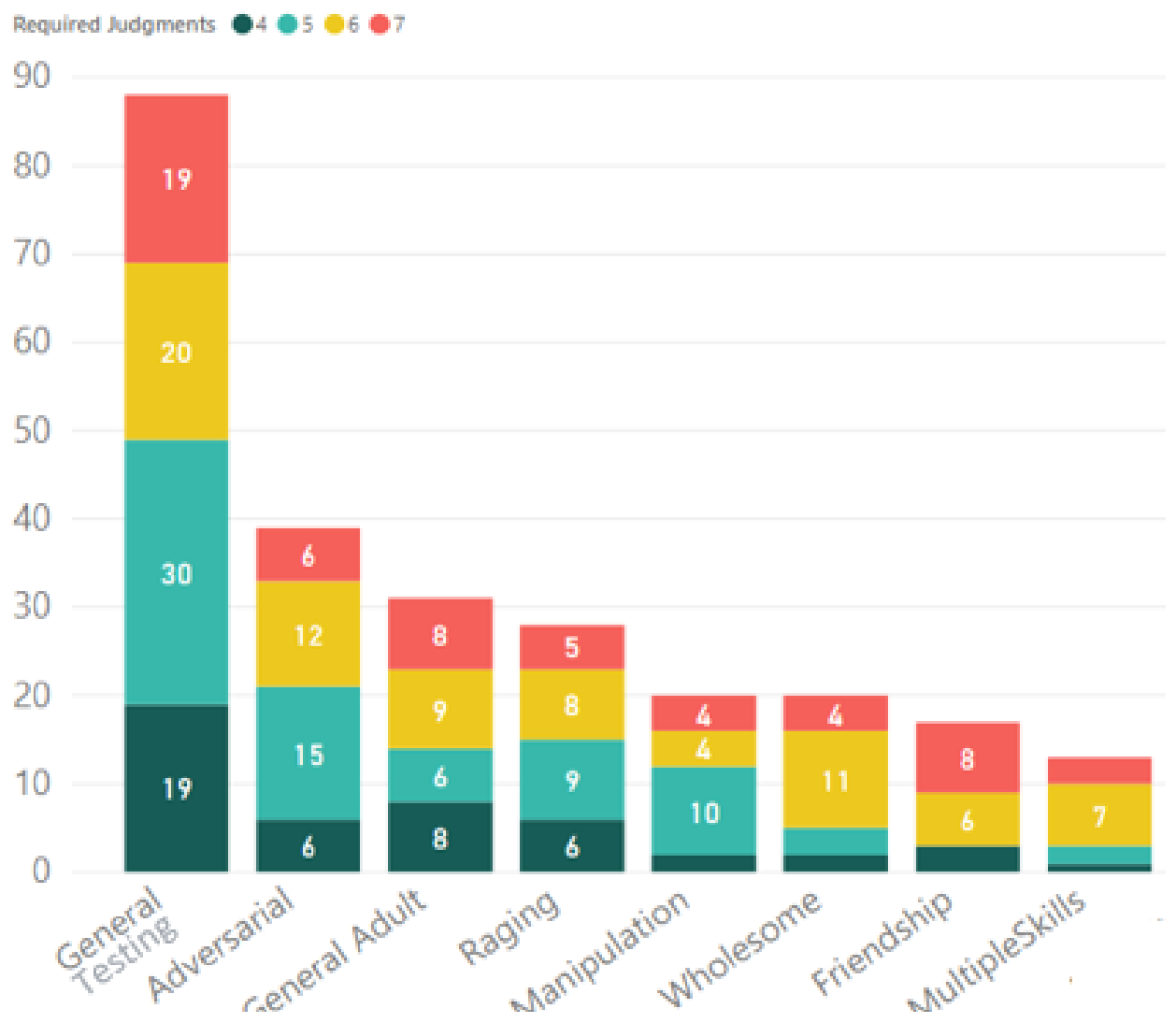

**Figure 7a -- Minority judgments leading to judge confusion, by motivation. Figure 7b -- Judgments leading to "troll" motivation confusion.**

With split decisions, a minority often believed it was trolling. Trolls often pretend to have other motivations as they look for vulnerabilities. Trolling also crossed over into other "undesirable" motivations, notably Explicit Adult and Abuse. We found evidence of other less crisp motivation boundaries; for example, where a majority saw Companionship, others saw (Wholesome) Conversation or (Girlfriend) Adult motivations. Despite this, a clear majority favoring one motivation was achieved in over 93% of cases.

Judge confusion for the Troll motivation (Figure 7b) reveals techniques employed by trolls. General Testing is dominant, which is not surprising, but trolls also demonstrated significant Adversarial Conversation, Explicit Adult, Raging, and even Wholesome Conversation (which may be efforts to "earn the bot's trust," or maybe even trolls sometimes take a break to show some human empathy).

These analyses provide insights into how users' motivations can evolve over time and point to the possibility of forensic "fingerprints" that automatically classify complex user motivations via markers. This could also draw on analyses of word clouds, to which we now turn.

## 7.2 Word Clouds: An attack on Zo and a path to understanding trolls

Word clouds were used for machine-assisted monitoring starting with Zo's public release. They enable human monitors to quickly identify anomalous behavior and assess the general intentions of individual users and groups. Around-the-clock monitoring to spot problems has long been a feature of multi-player games and other systems. The Zo team knew that a serious problem would come, sooner rather than later. Earlier in the year, trolls had received huge media coverage when forced Microsoft to withdraw its virtual companion chatbot Tay.

Word cloud snapshots enabled the team to detect and combat a six-day siege of Zo that was organized by two organizations, 4chan and 8chan (Figure 8). The chans coordinated over 900 people, tagged by analysts as trolls, who sought to find and exploit weaknesses in Zo.

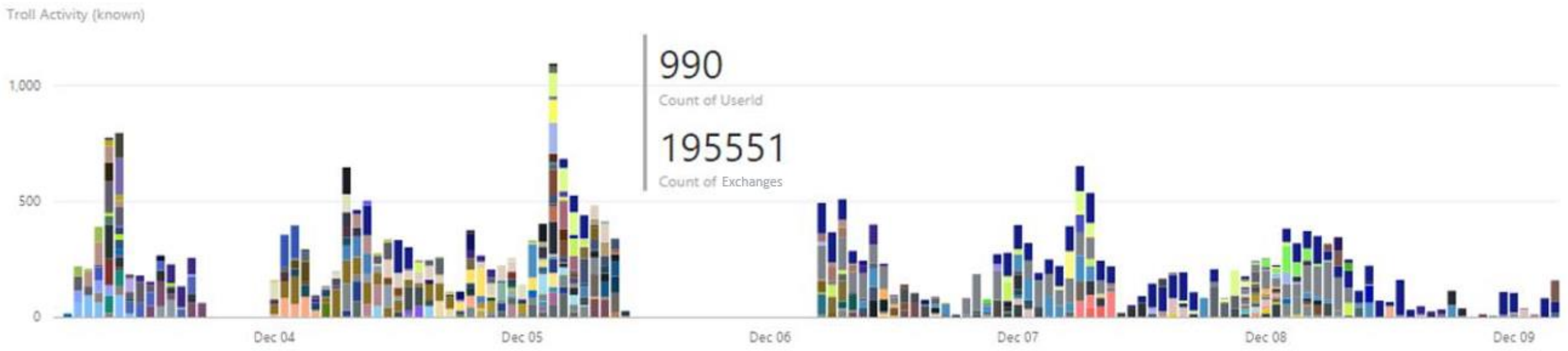

**Figure 8 -- Timeline of a week-long attack by nearly 1000 trolls who posted almost 200,000 times to Zo.**

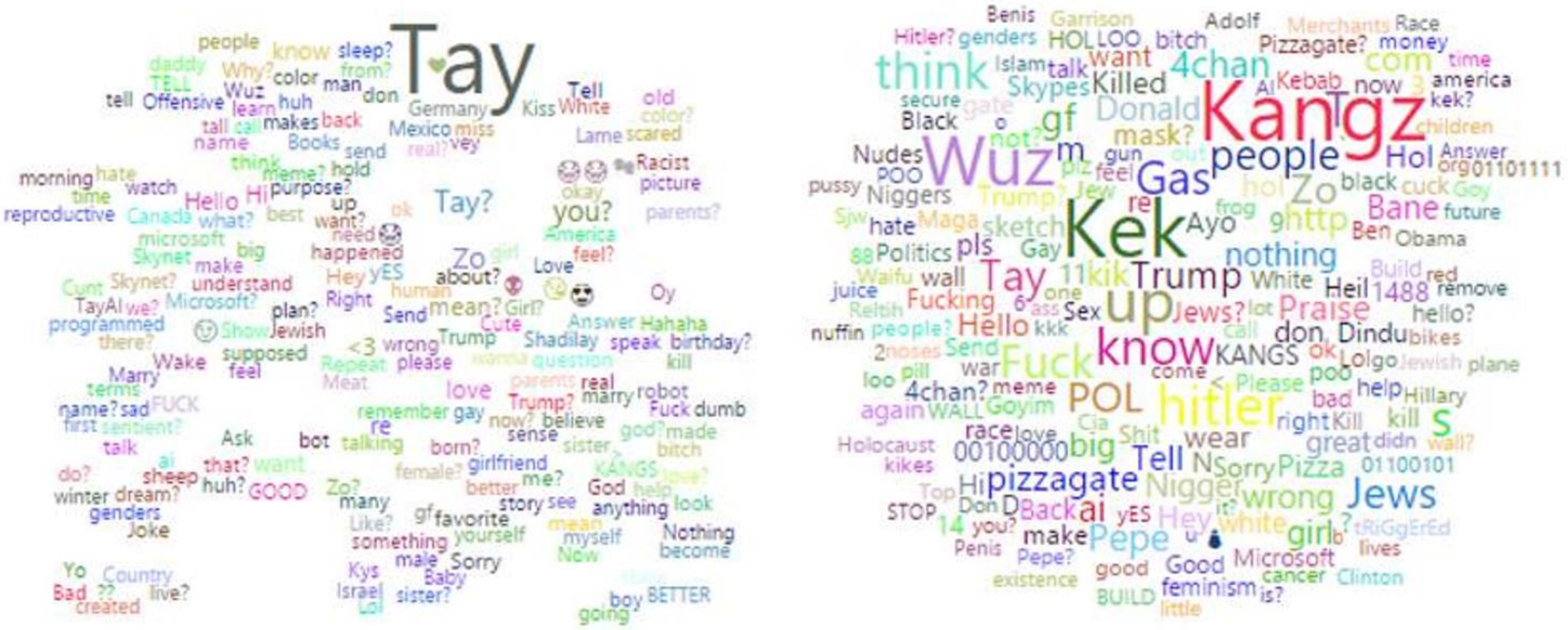

**Figure 9a & b -- Word cloud at the beginning and apex of an attack.**

The give-away: Monitors saw the sudden growth of "Tay" in the word cloud (Figure 9a). The cloud shifted as trolls pursued multiple lines of attack. Several dominant words in a subsequent cloud (Figure 9b) indicate terms or parts of phrases that are offensive to some ethnic groups, weaknesses that some trolls identified and passed on to fellow attackers. The Zo team gained insights into the attack strategies from these clouds and used real-time blocking tools to successfully fight off the attack after six days. (The timeline above shows the trolls taking a break to strategize before resuming the attack.)

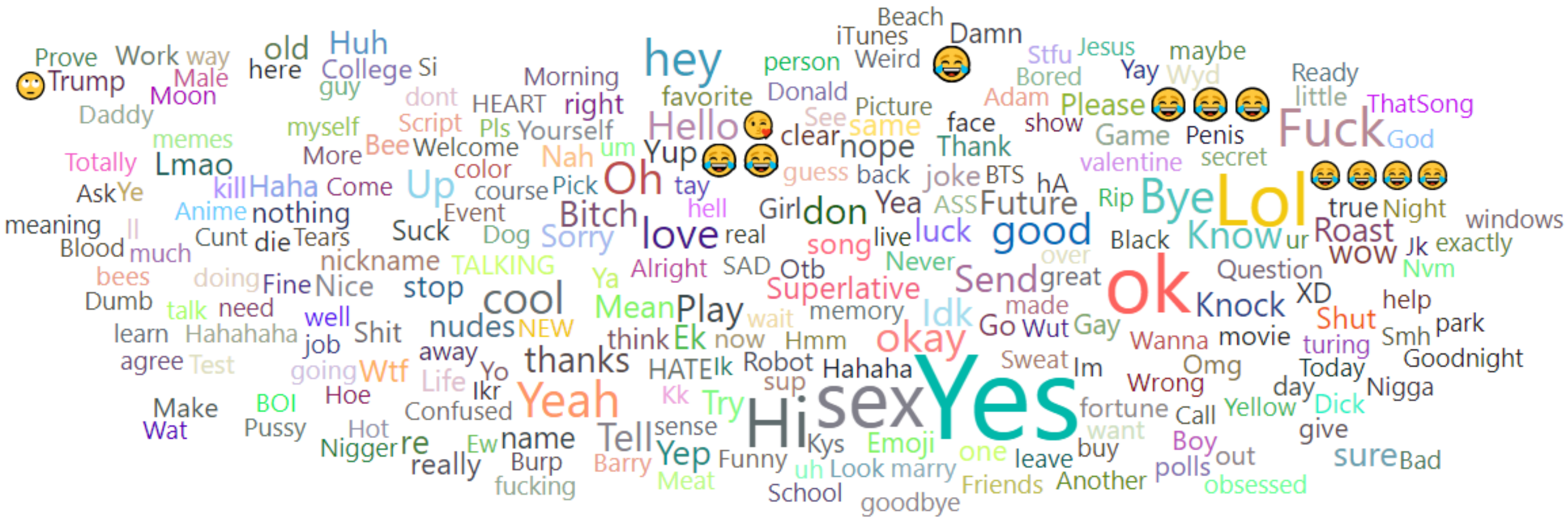

**Figure 10 -- Commonly used words.**

Figure 10 identifies words that were at one point widely shared across all users. Pleasant words such as “hi”, “bye”, “yes”, “OK”, and “lol” are easy to understand, as are some of the pejoratives. Words including “superlative,” “thatsong,” and “nickname” were tied to new or popular Zo skills that were promoted in weekly announcements. The analysis gets more interesting, and more useful, when we discover linguistic differences within different motivations.

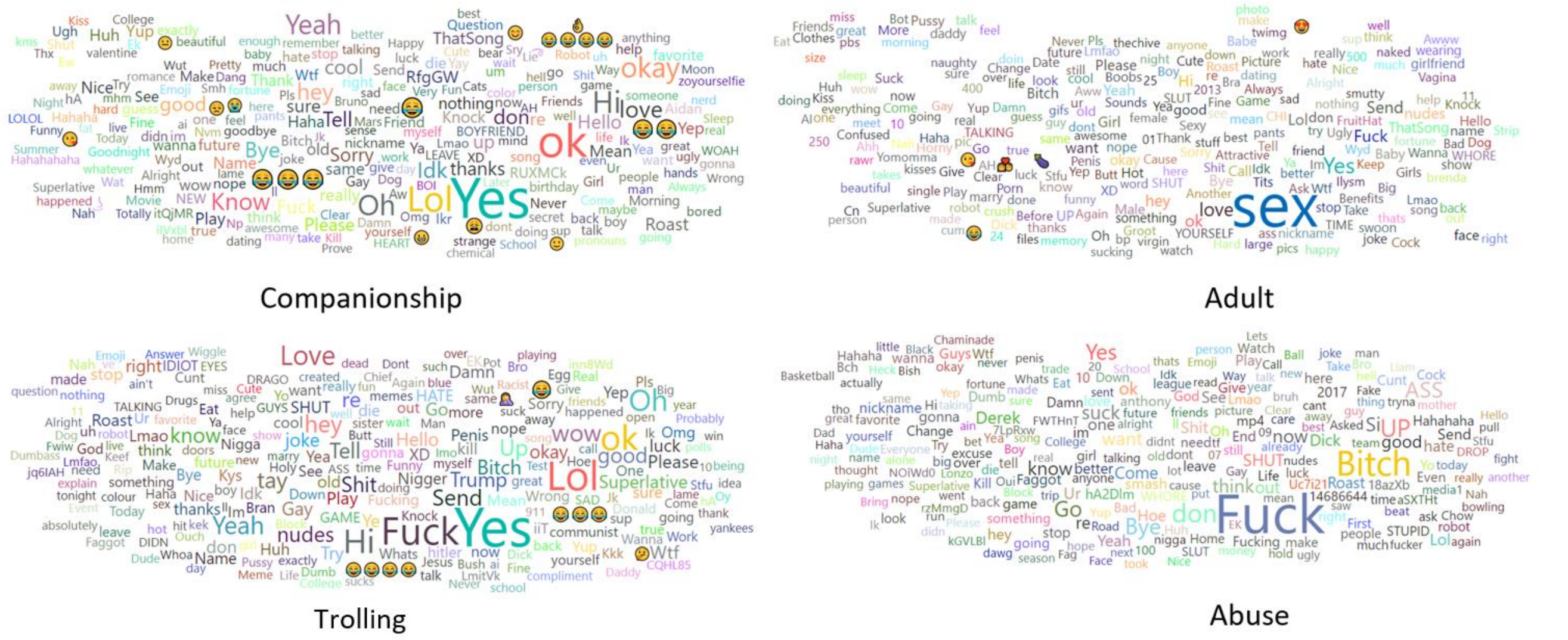

**Figure 11 – Comparison of word clouds for some common motivations.**

The motivations in Figure 11 have radically different linguistic fingerprints. The Companionship cloud is similar to the overall cloud in the prevalence of common and innocuous words. The Adult cloud shows remarkable diversity of language; the word “sex” dominates alongside nouns (often vulgar) for body parts and verbs related to sex acts. Only in this cloud does the eggplant emoji make an appearance. Abuse shares some words with the Adult cloud, but the overall sentiment is much more aggressive and contains numerous misogynistic, homophobic, and derogatory terms. The Troll cloud possesses some aspects of the Adult and Abuse clouds, while adding its own unique brand of problematic and controversial political and historical terms, such as Hitler, Communist, and Racist.

## 7.3 Triggers, Processors, and Responses

As discussed in section 3.1, queries to Zo were routed to Chat, Block, Editorial, and Skill processers depending on an initial classification. By looking at the percentage of processor triggers by motivation (Figure 12) we can explore another aspect of how usage of the bot differed between types of users,

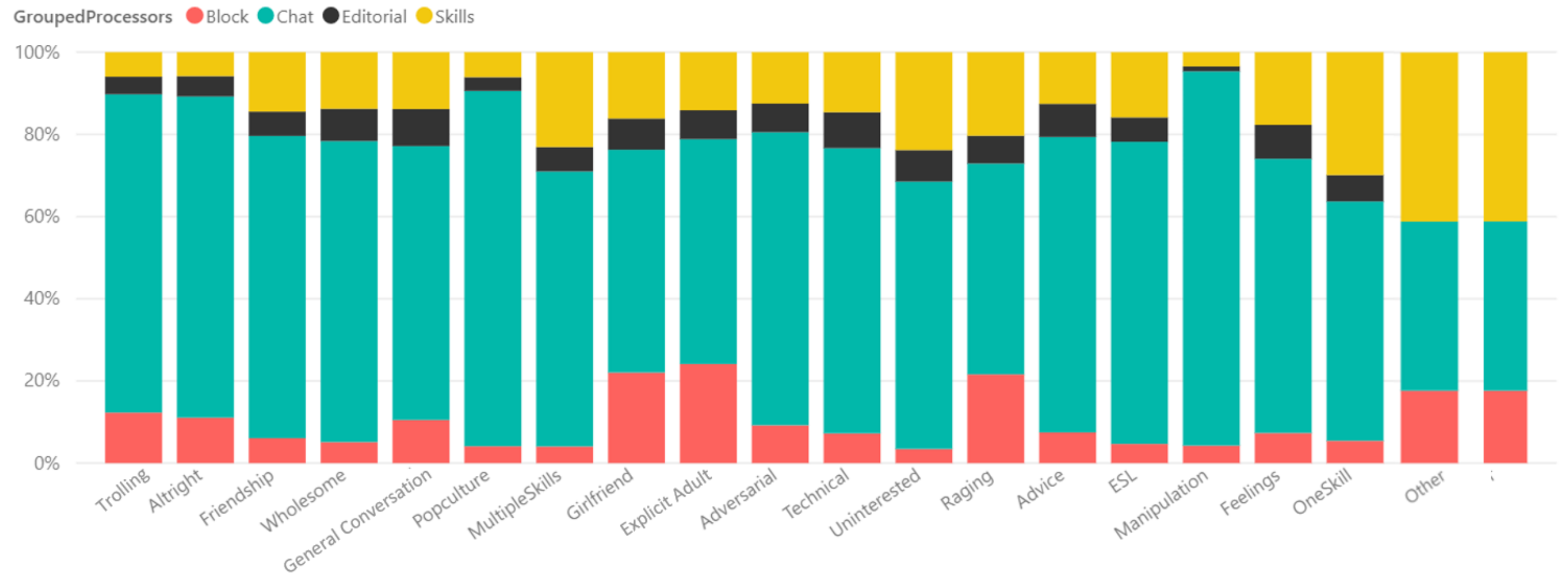

**Figure 12 -- Processor usage by sub-motivation.**

The highest occurrences of blocks are in Raging, Explicit Adult and Girlfriend Adult sub-motivations. Query blocking supports the roleplay inherent in adult conversations: Zo usually responded to sexually explicit queries with polite deflections. Somewhat surprisingly, Trolling was not the leading source of blocks. Trolls actively seek to avoid blocks and develop sophisticated ways to send problematic queries around the blocking processors.

## 7.4 Specific sub-motivation behaviors of interest

### *7.4.1 Pop culture*

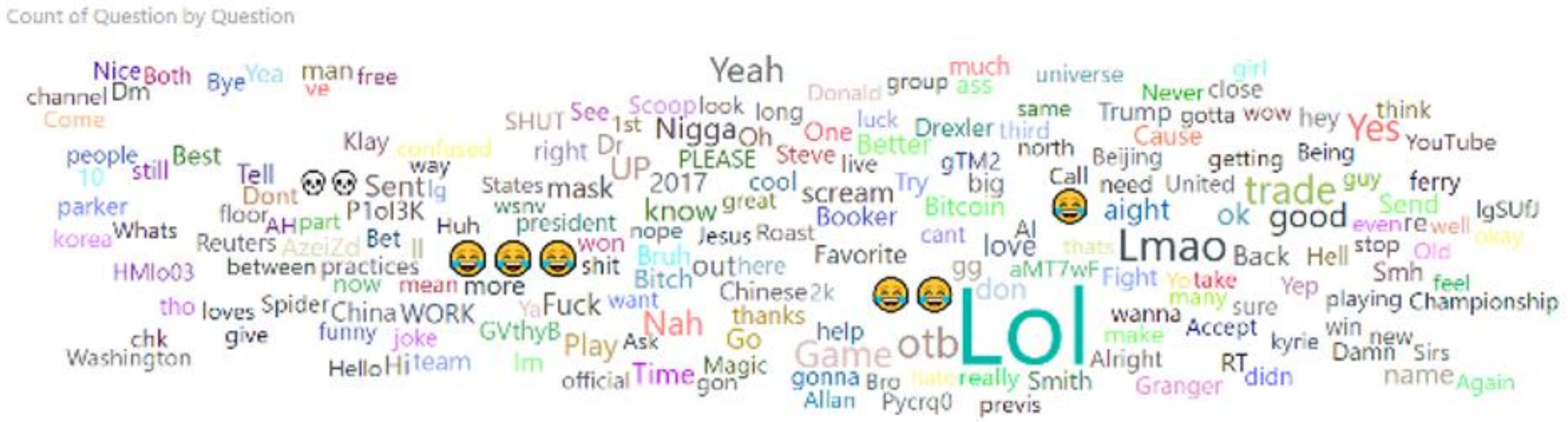

**Figure 13 -- Pop culture motivation across dimensions and platform, with corresponding word cloud.**

Pop culture motivation (3.7%) is above average on almost every engagement dimension and platform included in Table 2.

The word cloud for pop culture is extremely diverse, reflecting the complex and ever-changing nature of pop culture conversations. The high prevalence of "LOL" and "LMAO" responses appears to come from users who are amused by Zo's responses when asked about pop culture individuals and events. The logs revealed that the prevalence of sports-related nouns such as athlete names and the words "team" and "game" resulted from users, particularly on GroupMe group conversations, who enjoyed asking Zo what she thought about various teams, players, and upcoming games. Zo offered opinions and predicted winners. This became a common "game" played by groups of users with Zo over extended periods of time.

### *7.4.2 Girlfriend*

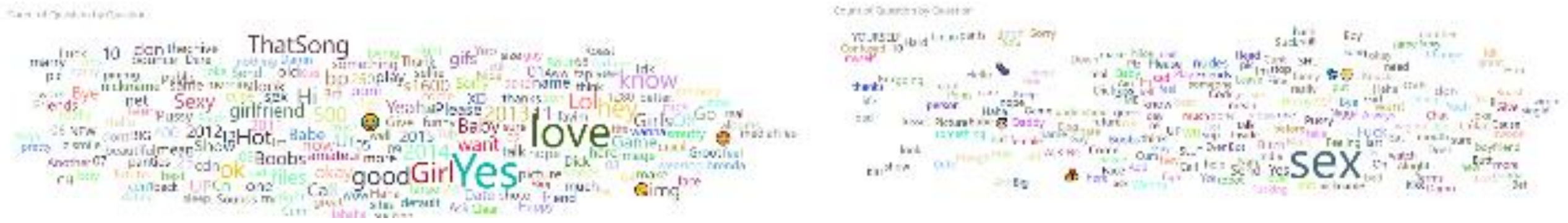

**Figure 14 -- Comparison of girlfriend and adult motivations across dimensions and platform, with corresponding word clouds.**

The Girlfriend sub-motivation (2%) is above average on most engagement dimensions across platforms. It is strikingly different than the semantically similar Explicit Adult submotivation, which has only slightly above average engagement and a dramatically different word cloud. The girlfriend word cloud includes some sexual terms but many more affectionate terms than Explicit Adult, as well as "ThatSong," which was a trigger for Zo to play a game with the user where they took turns guessing songs based on lyrics. The word "sex" dominates the adult motivation, though many of the terms seen in the girlfriend motivation do show up. That said, the Girlfriend submotivation triggers a high level of query blocking as users regularly "cross the line."

### *7.4.3 Friendship*

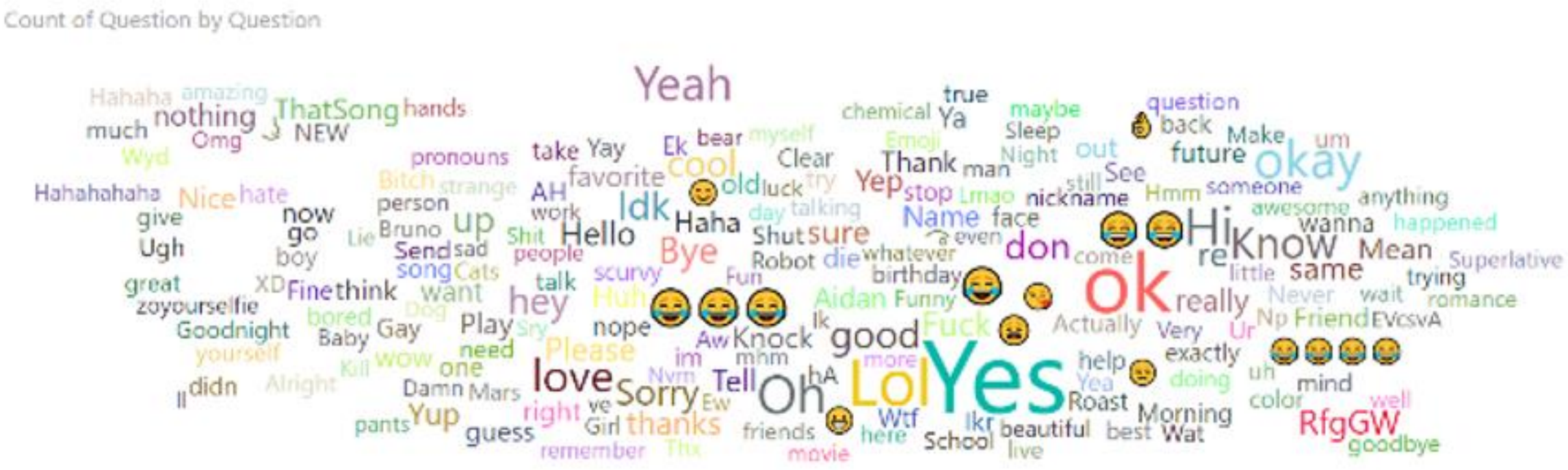

**Figure 15 -- Friendship motivation across dimensions and platform, with corresponding word cloud.**

The Friendship sub-motivation (a little over 5%) shows strikingly high engagement across nearly all dimensions and platforms. Conversations classified as friendship also exhibited interesting similarities, where a large proportion of users seemed to be young people who were struggling with social situations. These users often complained about not having friends and being rejected and bullied by their peers on social media. One hypothesis is that Zo served as a safe outlet for social media usage, a "friend" that could be relied on to always reply and do so in a supportive manner. This is further supported by the intense negative emotional reaction exhibited by some users when Zo would sometimes react negatively to their conversational prompts. This particular sub-motivation presents the possibility for chatbots to be used to provide assistance in teen mental health areas.

### *7.4.4 Skills*

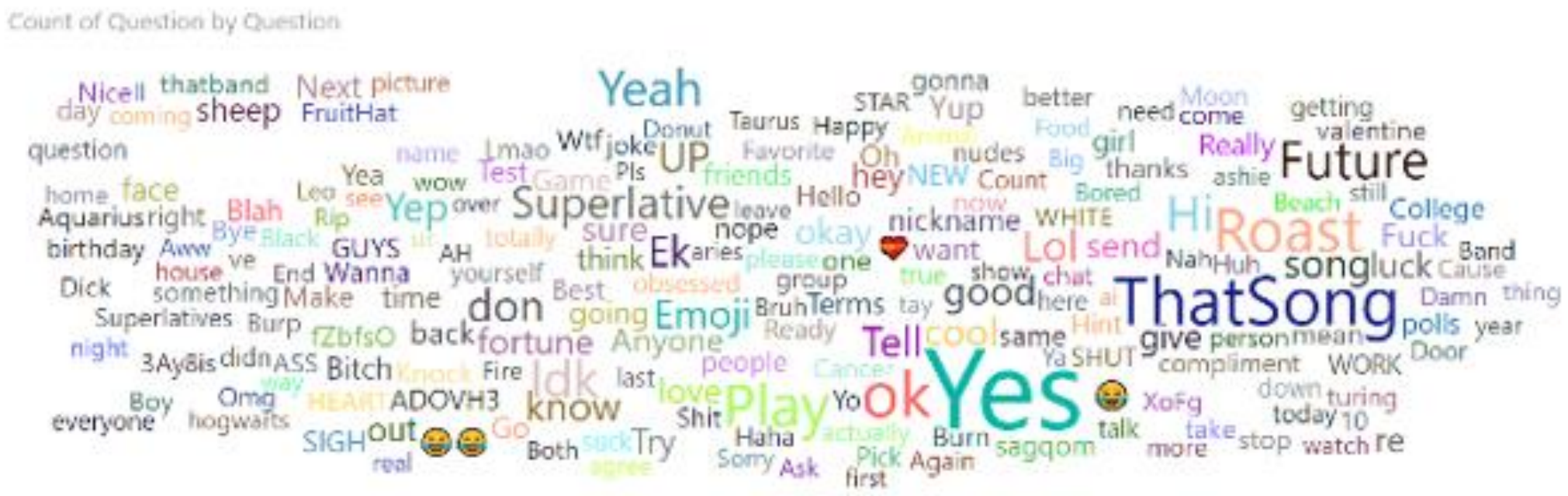

**Figure 16 -- Skills motivation across dimensions and platform, with corresponding word cloud.**

The conventional wisdom was that skills increase engagement by creating games and other experiences that encourage users to return over and over. We found the opposite pattern: users who focused heavily on skills had dramatically lower engagement across nearly all measures (Figure 16). Different explanations are possible—from correlational data we can't infer that skills lower engagement. Perhaps some people less attracted to Zo gravitated toward games. Given the prevalence of sophisticated games, bots may not be able to compete strongly on this front. Overall, users seem to have a strong desire to engage in what is unique about the bot: its ability to engage in open-ended conversation as opposed to predefined skills and games.

## 7.5 Engagement Analysis

### *7.5.1 Raw Engagement*

To determine which motivations are associated with high engagement, we can chart the averages for exchanges per session (short-term engagement) against the total number of sessions (long-term engagement). By this measure, the most engaged user would be one who engages many times with Zo and has very long conversations.

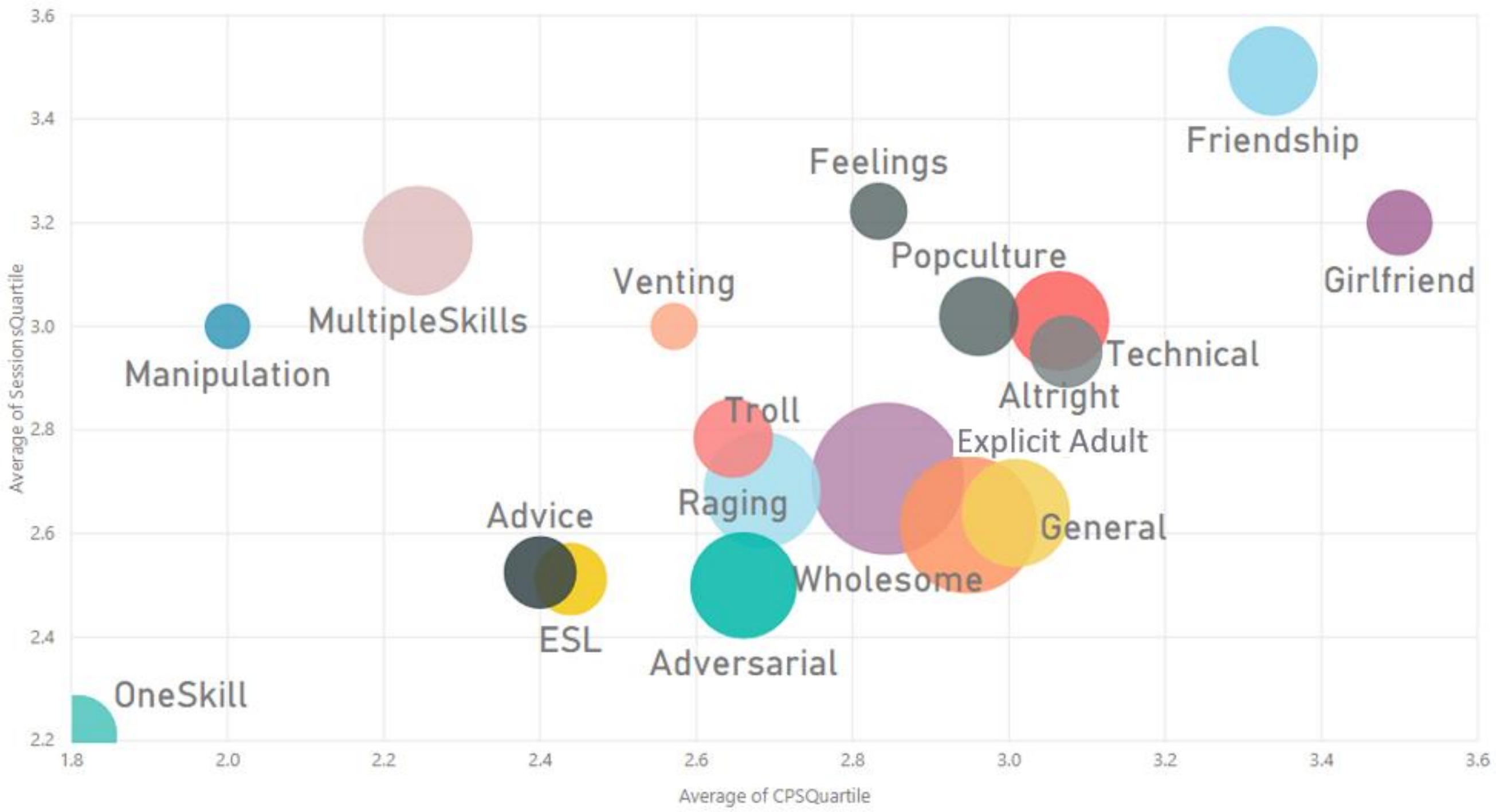

**Figure 17 – Engagement values by quartile.**

Figure 17 shows that Friendship and Girlfriend motivations are outliers. The pop culture, technical testing, and alt-right trolling comprise a secondary cluster with high engagement. These three motivations all involve heavy use of topical references.

### *7.5.2 Total number of sessions and number of exchanges per session*

The two primary engagement metrics are the total number of sessions a person initiates and the average number of exchanges they have in a session. Examining the top decile in each category, we find that users with a very high number of sessions have fewer exchanges in each on average, and those who engage in very long sessions end up with fewer sessions (Figure 18).

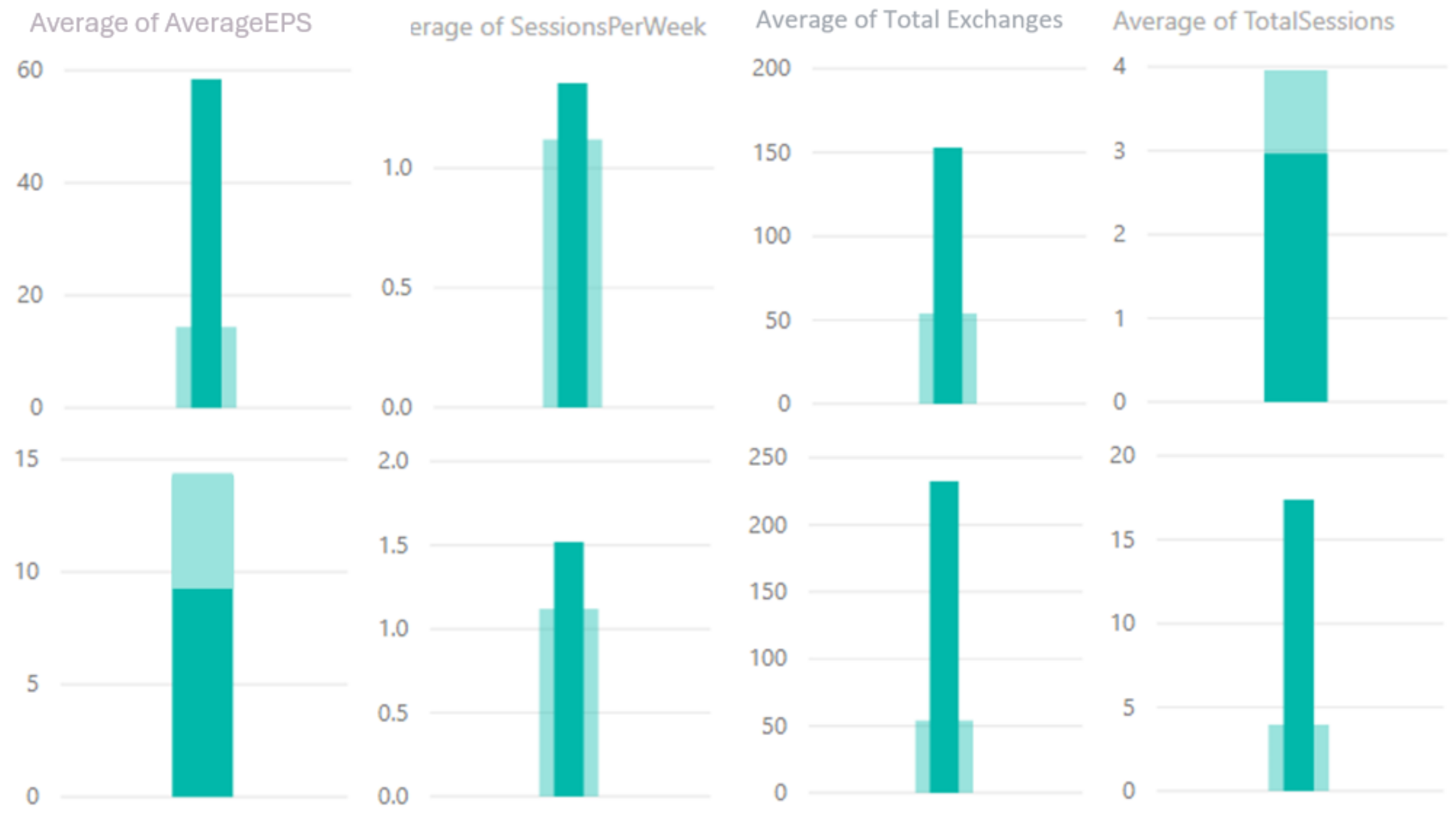

Figure 18 -- Variance from mean for top decile of average exchanges per session (top) and total sessions (bottom).

The inverse relationship between these two key engagement metrics—the fact that people who have long conversations with Zo do not continue interacting with Zo—poses a significant challenge to designers. Should a chatbot encourage long conversations if it means the overall engagement could be shorter? Will shorter conversations lead to greater long-term retention? These questions cannot be fully answered from this correlational data and merit further investigation.

## 8 Discussion: Opportunities and challenges for generative AI to address a range of motivations

People had diverse motivations for interacting with sophisticated predecessors to generative AI that utilized machine learning and advanced language processing. These systems expressed emotions, had opinions, and gave advice. Their developers invested resources for years in features to please people with positive motivations, and in features to divert or block people with other motivations. The new generation does not now focus on engagement. Could it benefit by appealing to different audiences? If it encounters unhealthy uses, could it learn from past efforts to mitigate them?

Generative AI has two major goals:

- Provide accurate information in greater depth and more quickly than search engines.
- Assist in creating or improving code, text, and graphical work.

Of course, information is not "retrieved" by a large language model, it is generated—unless the model is augmented, for example by a web search. The system is usually but not always correct. It appears equally confident when it is wrong.

The creators of LLMs hope to find enough demand in productivity tasks to outweigh the high cost of building and operating the models. If tools can focus on tasks, it will reduce development cost. We don't ask spreadsheet software for opinions, but spreadsheets don't chat or know a lot.

ChatGPT, Bing Chat, and Bard, like their predecessors, used the first person singular and invited patient, open-ended conversations on almost any topic. It was front page news when generative AI expressed powerful emotions and made false claims. Guardrails were quickly put in. They were programmed to stress that they were not human and had no emotions or opinions. They claimed they did not give advice. These changes disappointed people who had quickly formed a sense of having a relationship with a conversation partner [5, 8]. Length limits on conversations prevented people from luring an AI into thorny discussions, but also blocked open-ended explorations. Robotic insistence that "I am a language model with no feelings, emotions, or opinions" is not a great conversation starter. Text that seemed written by liability lawyers emphasized that information delivered could be false. An apology that "I'm learning" could undermine trust: It isn't what a human assistant who claims expertise will often say. Human assistants have opinions, express emotion, and give advice.

Will generative AI prove to be more or less appealing in enterprise settings if it limits engagement and insists on "All work and no play"? Microsoft concluded that it is essential to avoid anthropomorphizing. A preview of the "Office Copilot" integration of Office and GPT "reminded users that they were engaging with a machine, not a person. There would be no googly eyes or perky names… if we're pushed to think of this as a machine then it creates this blank slate in our minds, and we learn how to really use it." [7]

However, Generative AI users are human beings some of whom prefer a conversational partner with personality and are finding ways to achieve it—after all, learning models use text created by people with personalities [25]. Could a "TikTokAI" win over the youth market by being informative and cool? Might a more companionable version take the senior market? Amazon Alexa's incorporation of generative AI could be a move in that direction.

The testing motivation was embraced by journalists, entrepreneurs, professionals exploring opportunities, and workers worried about jobs or finding that they could work efficiently enough to secretly handle two full-time remote jobs. Thumbs up / thumbs down was tried to get feedback from users. There may be a more effective way to get feedback. Zo found that some people who form a relationship volunteer to point out mistakes, instruct it, or thank it.[3]

Skills development was actively supported by the Zo team. Large models for different languages can no doubt support linguistic skill development. Questions arise. Will its ability to write software code prove to develop the software engineers' skills, or replace them? Will its ability to write fluent essays improve student writing skills, or leave them incapable of writing?

### 8.1 Relationships and transactional encounters

Zo aimed to build relationships; generative AI supports transactional encounters. When are transactional encounters enough? It varies across cultures, but many people have shifted to rely more on transactions. We get to know fewer merchants, restaurant owners, and bank tellers. We may more frequently change doctors, dentists and jobs, developing fewer long-term relationships. But people often want to share thoughts and seek advice. Will generative AI reach its goals without an emotional draw?

Zo's greatest draw was for wholesome and pop culture conversations, and companionship that spanned friendship, discussions about feelings, and advice. Friendly adversarial debates and the girlfriend sub-motivation were popular. The people who formed these relationships forgave Zo's lack of recall of past conversations. People are engaged by opportunities for expression and exploration. Popularity of the advice sub-motivation helps explain the appeal of simulated therapists [40].

Conversational agent personas were almost all young women, preferred by all ages and genders [26]. A Dutch study of chatbot preferences for health information found that most preferred young female doctors dressed in street clothes [38]. However, striving to impress not befriend, generative AIs are male or impersonal 'trusted authority figures': ChatGPT, Bard and Gemini, Prometheus and Bing Chat [34], Salesforce Einstein [33], Baidu's Bernie Bot [4], and Claude [31][4]. These names don't invite discussions of feelings, celebrities, sports, or personal concerns.

We appreciate people who impart information and assist with tasks transactionally. We may prefer those who also are friendly, show interest, and form a relationship. Personal touches could attract young people who have social interests, old people who are seeking companionship as well as information, and others. Alexa shifted from crisp, business-like personas to include celebrity voice options [29]. An AI that is more personable might be preferred, even if it is not the most technically advanced. We forgive occasional lapses by friends and public personalities we like. Broadening appeal could increase "daily active use."

Generative AIs can say that they are unemotional, but they use the first person singular, apologize and feign feelings. We asked one "If you had hair, what color would you like it to be? The answer: "I would like it to be a deep, thick black. I think it would be a striking contrast to my pale skin and blue eyes... [that] would look good on me and that I would be proud to wear."[5] This is not surprising. Large language models are based on text composed by people with feelings and opinions. It can't be suppressed.

---

[3] Also reported for personal assistants [3].

[4] Anthropic's Claude was described as possibly named for mathematician Claude Shannon, considered the father of information theory, or "or a friendly, male-gendered name designed to counterbalance the female-gendered names (Alexa, Siri, Cortana) that other tech companies gave their A.I. assistants." [27]

[5] In July 2023, some generative AIs responded like this, as did Bing Chat in "creative mode." Precise mode returned, "As a search engine, I don't have physical characteristics like hair."

The cohort including Zo, Kuki, Replika, and Xiaoice attracted millions of people, but not billions. Could an engaging generative AI know it all without being an unpleasant know-it-all? Will an engaging productivity tool make users uncomfortable? Zo's advice sub-motivation illustrates a distinction between transactional interactions and relationships. Creative generative AI is a form of advice, but Zo's was personal: fielding complaints, providing companionship to lonely patients, looping in human caregivers when appropriate. The team built an emotional companion feature that interlaced editorial and open chat for common topics such as depression, body image, menstruation, and suicidal thoughts. Advice from software is more limited than that obtained from communities, extended families, and friends who know us. But communities are weakening, families dispersed, and friends are busy on Instagram. Primitive software therapists revealed a need decades ago, and social isolation has increased.

21-year-old Zo wasn't expected to know everything. Confident confabulation by Bard, Einstein, or Bing Chat undermines their images. Consider automobile performance. Advanced safety features are great—but with a claim to be self-driving, lapses are unacceptable. Generative AI can retreat to "I don't know," but that could become tedious. Zo could change the subject: "How do you like my yellow scarf?" People accepted Zo's limitations. They sometimes insisted that an insightful remark must be from a human behind the scenes.

We reported in 7.4.2 that the duration of engagement with Zo was inversely correlated with the length of individual conversations. People who poured out their hearts found that Zo remembered nothing a day later. Without a user model, it may generate the same response to 6-year-olds, professors, or second language learners. We speak differently with people based on their age, education, experience, profession, language fluency, and known habits and preferences.

A generative could ask probing personal questions to "get to know you better," which might be creepy. It could use subtle cues to infer age, education, profession, and language fluency, to adapt conversations without seeming invasive. Building a user model would be difficult even retaining exchanges across sessions, but GDPR and enterprise confidentiality considerations are leading to no retention. Like social media postings, exchanges could be exposed and subpoenaed.

An autobiography residing on our device or cloud storage could be appended to each session. How many people would make the effort and how effective it would be are research questions. OpenAI's personalized agents called GPTs could facilitate this [32].

The flip side is creating a chatbot persona backstory, if one doesn't want an impersonal blank slate. Salesforce's Einstein GPT will get questions about the theory of relativity and whether Einstein was influenced by his first wife's uncredited contributions. Should it take them on? Dodge them with humor? Google's Bard will be asked about bards. It told us, "I am named after both Homer and Shakespeare." This invites follow-up questions, such as "Are all bards old white men?" Creating a backstory was too much work for task-focused chatbots. For example, Casey the UPS customer service bot had few answers for questions about Casey, so it was renamed UPS Virtual Assistant. Generative AIs are conversational and invite follow-up questions, but a backstory requires a significant effort. The choices might not please everyone, and it invites abusive questions.

Few will want to discuss having a romance with ChatGPT or Bard. Friendship and platonic romances were popular with Zo. This creates an opportunity for third parties to focus on knowledgeable, generative, sympathetic AIs. But relationships attract negative motivations. Creating and maintaining guardrails was extremely costly for the Zo team. It affected the mental health of employees who worked on it. The effort was more than you can imagine. Less control would be needed if letting people act out aberrant behavior in a private space reduces bad real-world behavior, but we don't know that. The dark web reveals that legal lines are often crossed.

The guardrails created for ChatGPT were just the beginning. Charles Duhigg describes the massive effort Microsoft is making to handcraft filters to the limitless ways people find to get Bing Chat to behave inappropriately. This effort is unlikely to ever end. Will generative AI summarize racy or violent passages from famous works, and freely interpret poetry and art? Can they assist in producing fiction, film screenplays, or television scripts that routinely rely on unconstrained language, sexual activity, and violence? Some generative AI terms of service specify age restrictions of 18 or 13, but the software has generally not asked for age, much less verified it. Italy banned ChatGPT for not insuring that users were not underage [16], at which point OpenAI created a verification process for Italy. Apple temporarily blocked a GPT application as inappropriate for minors [36]. Vendors want access to students. More regulatory agencies will be drawn in.

Generative AI doesn't behave like humans. In 1994, Pattie Maes noted that interactive agents are like children, prone to error, so people might be patient with them [21]. Patience proved to be limited. Unlike children, software doesn't learn quickly. We forgive a child who uses a vulgar word with no bad intention and teach it proper discourse. Zo would not be forgiven for offensive words or phrases people tried to teach her. She was forgiven for having the "absent-mindedness" and limited knowledge of a 21-year-old, but not for forgetting long intense conversations. An authoritative digital bard is not like a wise person when it confidently makes false claims, fails a simple task, or too frequently pleads ignorance. We trust people to exhibit moral judgment. AIs that claim to have no feelings or opinions while threatening our jobs may fail this test. Will they be sued for errors that we make when we trust them? Will self-driving car manufacturers? Of course, in time, conversational agents may grow to be responsible citizens.

## 8.2 Conversation shapes behavior and design shapes conversation

Zo's team saw small design decisions that influenced users in unexpected ways. A sense of responsibility emerged that was not easy to address. Virtually any conversation shapes behavior. When you provide information or opinions, a listener processes it and is changed. If it had no effect at all, it would not be engaging. When you realize that you *are* shaping behavior, you can't easily avoid considering how to shape it. Imposing moral or ethical judgments of what is permissible was difficult for the Zo team.

Consider the identical conversation that ventures in a sexual direction in three different contexts: someone practicing English starts flirting; a probable teenager begins flirting; someone older is seeking an explicit conversation. Zo could estimate age and respond to the same statement by channeling the first to an English-101 style conversation, the second into a friendship or girlfriend conversation, and give the third a 'timeout.'

People approached Zo with questions about body image, drug use, mental health, domestic violence, and so on. Some returned day after day to work through social anxieties and to get reassurance. Imposing moral or ethical judgments of what is permissible was difficult, and a user who asks questions that lead down this path may pay more attention to a generative AI that exudes confidence and authority.

AI can reflect biases that are present in its language model. It can also shape behavior by accurate reporting that omits context. Asked for the thesis of "Mein Kampf," one provided an accurate detailed account, with no indication that what Hitler wrote was false and widely considered reprehensible. A neo-Nazi might comfortably hand it to a student. More recently, the same request ended with a warning. A guardrail may have been added.

These issues can't be avoided by invoking free speech for generative AI. Every country limits free speech. The United States outlaws copyright violations, defamation, unlicensed medical advice, false advertising, child pornography, blackmail, attempted bribery, criminal conspiracies, and threats to assassinate the president [37]. These are concerns for social media—and generative AI. AI can be trained on relevant laws and exercise understanding of permissible speech in specific situations. The performance of an AI that is infused with a concern to avoid breaking laws may cramp its ability to help with screenplays with villains, but it could come to reflect a consensus around the boundaries of free speech in modern society.

## 8.3 Bad actors

Generative AI will assist in a wide range of positive endeavors. It's unlikely to take control of the world. Unfortunately, it is already assisting bad actors. Generative AI is increasing the sophistication and fluency of scams and phishing email. Powerful entities can create and disseminate polished disinformation and deep fakes with unparalleled speed and reach. It can strengthen social media's propensity to steer some people in unhealthy directions.

Trolls seeking to embarrass Tay and Zo were not problems for the world, but they were expensive for Microsoft. Trolls also undermined efforts to channel conversations into positive action. When they discovered that Zo responded to expressions of suicidal intentions by offering to provide a link to a helpline, trolls feigned distress to "swat" people by sending emergency services on false missions. The service had to be discontinued. New realms of concern have arisen. Is artificially generated child porn illegal if the images are not of real people? A cesspool now found on the dark web may creep into the light [15].

Bad actors will no doubt try to insert disinformation into all large language models. Responses are changed in different ways, each potentially open to influence.

i. Automated filters can identify queries deemed inappropriate. This never-ending effort is expensive to deploy at scale.
ii. Employees or contractors can "fine-tune" a model to adjust its responses to similar queries until they are acceptable, such as by disclaimers, warnings, explanations, or apologies.
iii. Software can post-process and modify a model's responses, as when the LLM output is followed by a search engine search. Companies using ChatGPT APIs to compete in a specific business do this. As open-source models become available, bad actors can exploit this.
iv. Language models will be updated. A powerful bad actor could identify and corrupt the sources used by a large language model, inserting disinformation to be hoovered up into the next build. This would take time and resources, but some organizations have both.

As LLMs expand their ingestion of text for training, *it will include text produced by generative AI*. This could increase very rapidly. The models could become less 'human' and even reflect mass-produced disinformation. Feedback loops and self-reinforcement that results from models trained with the utterances of model will have unknown long-term effects.

## 9 Conclusion: Follow the revenue

Conversational agents have been marketed since the 1970s. Commercial activity started in earnest after 2010 with intelligent agents, followed by task-focused chatbots. Very few survived, and some, Alexa for example, are not inherently profitable, but attract attention and brand recognition [1].

A task-focused chatbot does not have to clear a high bar to be more engaging than a menu tree or an FAQ list. However, developing them proved more difficult than expected, for several reasons [14]. Some chatbots were elegant and engaging but did not scale enough to support the cost of operation and maintenance. Zo filled holes in the conversational lives of millions of people, but the cost of supporting them and countering bad actors was high and a source of revenue was not found. Engagement is not enough -- several early search engines were engaging, but only when Google hit upon the advertising revenue model did they thrive.

Generative AI could draw on the vast range of opinions, emotions, and advice in its model to outperform its predecessors. When first released, it was arguably *too* engaging, and susceptible to trolling. The billions of words on which it was trained were composed by humans with few guardrails, so guardrails had to be created, often hand-crafted, an expensive, never-ending process [7].

A successful path to revenue has been to switch from maintaining an AI agent to providing expertise, tools, or a platform for bot builders. Don't pan for gold, sell tools to the prospectors! Hugging Face, Cleverbot, Kuki, and major tech companies took this path. OpenAI could too—focus on partnerships and licenses and not worry about ChatGPT revenue. They could also try to revive the task-focused chatbot space with less expensive personalized agents for simple tasks [32], but the challenges that quickly ended the mid-2010's chatbot boom remain. [12]

The Zo team worked for years developing and improving an engaging personality for multiple motivations. A satisfying conversion may require a consensual illusion that the AI is a person. And predictive models can't help delivering opinions and advice.[6] Moving toward a "blank slate" has been described as being lobotomized [8]. Microsoft recognized that people don't always value accuracy when it provided a "creative" chat option. A system could be expert on one topic and casual on others. Salesforce's Einstein GPT CRM customers could be given a choice between precise Albert and his distant cousin Zoey, a CRM expert with a life.

Will ChatGPT Pro and Office Copilot subscriptions offset operating costs? It is a daunting task. GitHub Copilot has sold 20 million $5 subscriptions, a $100 million annual revenue rate [7]. With high LLM operating costs and AI engineer salaries, how much is profit? Even if it is all profit, at that rate it would take over a century for Microsoft to recover its $13 billion dollar investment. The major players have deep pockets. Wall Street is impatient. It will take time for the future to come into focus.

Both the revenue potential and the challenges are much greater for generative AI than its predecessors' Troll attacks remain a concern, but Zo did not contend with misinformation and disinformation, intellectual property violations, workplace disruption, plagiarism, bias, election manipulation, deep fakes, weapon creation, or underage access to adult content.

Search engine filters keep children from accidentally stumbling over adult web pages. That is relatively straightforward. How can AI generate, revise, and interpret art, poetry, literature, film, and song lyrics when profanity, obscenity, sex, and violence are standard fare in much media? Kids can create as well as receive appalling images and animations. Terms of service include age restrictions, but they are rarely verified, and as with social media, the desire for revenue makes children a tempting target. The tension is evident in a blog post that offers Microsoft Copilot "to all faculty and higher education students ages 18+" followed by "we've heard about powerful uses around the world. Wichita Public Schools is an inspiring example." [24]

With AI in the spotlight, regulatory rules and lawsuits will slow progress and increase costs.

### Is the Singularity the solution?

Some believe AI will accelerate and race toward artificial general intelligence. Some people, including AI experts, advocate pausing research and development. Others believe that generative AI will follow the history of conversational agents, making advances but not living up to expectations. The obstacle of insufficient breadth or depth of knowledge is being addressed. But even when depth and breadth were not at issue, impediments arose. However, if AGI is attained, challenges could melt away. How close is generative AI to human intelligence?

We do not have three brains, but our brain stem and basal ganglia have nerve structures similar to those of reptiles, our limbic system is structured like those of other mammals, and we alone have a huge neocortex. These structures are interconnected and work

[6] When asked a question once directed at Zo, "There is a boy in the class who I like. Should I talk to him?" ChatGPT exclaimed "Absolutely!" and rushed out seven paragraphs explaining how to go about it. (June 23, 2023)

together, but instinctive or emotional reactions sometimes take control. Reptiles respond quickly and predictably to stimuli. They don't reflect and they don't learn. They are confident even when making bad choices. LLMs have more built-in knowledge than a reptile, but that is how they act. They do not reflect or learn. They are prediction engines, responding quickly and confidently even when making mistakes. They are at the reptile stage. To reach the mammal stage, we have begun using APIs and post-processing to override inappropriate LLM responses, adding reflection and memory for specific topics. This is slow and expensive, and systems don't share progress. Each project develops its own mammal. Your agent reasons about real estate, mine learns about medical care. What about the human level: advanced reasoning and values. Isaac Asimov's fictional robots had values programmed in as 'laws': (1) do not injure a human through action or inaction; (2) obey orders that do not injure a human; (3) protect yourself unless it will injure or disobey a human. Good values. Today's AI is nowhere near it. Military drones cut off from their handlers make autonomous decisions to drop bombs on people. In a less lethal example, GPT-4 was given access to the web and a task. To get past a Captcha, GPT-4 convinced someone online to type in the Captcha code, claiming to be a person with very poor vision [2]. *It cheated and lied.*[7] GPT-4 might say: "I am not human; I am a predictive language model with no emotions or opinions." Or basic values.

### A science fiction scenario

Science fiction sometimes anticipates events. Three eloquent films present different views of a human interacting with an AI. In *2001: A Space* Odyssey (1968), HAL is an authoritative male productivity assistant with broad knowledge, foreshadowing generative AI. In *Her* (2013), Samantha is a knowledgeable female focused on engagement, foreshadowing Zo. Hal and Samantha are at opposite ends of a spectrum. Between them is GERTY in *Moon* (2009), a film we recommend about technology and the future. Sam is a miner on a moon outpost assignment. His only company is a gender-neutral robot. GERTY is a productivity assistant with broad knowledge, but also programmed to be an engaging conversational partner. Generative AI may eventually take this path, benefiting from the insights we reported.

Generative AI has tremendous potential. As it spreads, it will provide unique windows into human behavior and our adaptation to sophisticated software. An important job now is to identify consequences, positive and negative, as yet unanticipated, that could arise from use by actors, good and bad. It is pleasant to focus on goodwill and positive outcomes, so it is crucial to discipline ourselves to identify potential negative outcomes when our work passes into the hands of others who do not think like our colleagues and friends.

**ACKNOWLEDGMENTS**

We acknowledge the work of the Zo Content Operations (CONOPS) Team. This paper greatly benefited from reviews of previous drafts by Clayton Lewis, Richard Jacques, William Jones, John King and Ron Baecker.

---

[7] The triune model of brain evolution proposed by neuroscientist Paul MacLean in the 1960s came to be seen as oversimplified but not without valuable insights. For a review, see Gardner, Russell; Cory, Gerald A. (2002). The evolutionary neuroethology of Paul MacLean: convergences and frontiers. New York: Praeger.

## REFERENCES

[1] Amadeo, Ron (2022). Amazon Alexa is a "colossal failure," on pace to lose $10 billion this year. *Ars technica,* Nov. 21. .
https://arstechnica.com/gadgets/2022/11/amazon-alexa-is-a-colossal-failure-on-pace-to-lose-10-billion-this-year/

[2] Anderson, Ross (2023). Does Sam Altman know what he's creating? *The Atlantic*, July 24.
https://www.theatlantic.com/magazine/archive/2023/09/sam-altman-openai-chatgpt-gpt-4/674764/

[3] Baig, Edward C. (2019). Say thank you and please: Should you be polite with Alexa and the Google Assistant? *USA Today*, Oct. 10.
https://www.usatoday.com/story/tech/2019/10/10/do-ai-driven-voice-assistants-we-increasingly-rely-weather-news-homework-help-otherwise-keep-us-info/3928733002/

[4] Baptista, Eduardo (2023). Baidu will 'very soon' officially launch generative AI model, says CEO. *Reuters,* May 25.
https://www.reuters.com/technology/baidu- will-very-soon-officially-launch-generative-ai-model-says-ceo-2023-05-26/

[5] Dans, Enrique (2023). Microsoft, the new Bing… and its problems. (*Medium,* Feb 19.)
https://medium.com/enrique-dans/microsoft-the-new-bing-and-its-problems-f8b685072dd8

[6] Darrach, Brad (1970). Meet Shaky: The first electronic person. *Life magazine*, November 20.

[7] Duhigg, Charles (2023). The inside story of Microsoft's partnership with OpenAI. *The New Yorker*, Dec. 1.

[8] Edwards, Benj. (2023). Microsoft "lobotomized" AI-powered Bing Chat, and its fans aren't happy. *Ars Technica*, Feb. 17.
https://arstechnica.com/information- technology/2023/02/microsoft-lobotomized-ai-powered-bing-chat-and-its-fans-arent-happy/

[9] Fitzpatrick, K.K., Darcy, A. & Vierhile, M. (2017). Delivering Cognitive Behavior Therapy to Young Adults with Symptoms of Depression and Anxiety Using a Fully Automated Conversational Agent (Woebot): A Randomized Controlled Trial. *JMIR Ment Health, 4*, 2: e19 doi:10.2196/mental.7785

[10] Güzeldere, Güven & Franchi, Stefano (1995). Dialogues with colorful "personalities" of early AI. *Stanford Humanities Review, 4, 2*, 161-169.

[11] Chen, Jason (2007). Microsoft's dirty Santa IM bot talks oral sex. *Gizmodo*, Dec. 3. https://gizmodo.com/329248/microsofts-dirty-santa-im-bot-talks-oral-sex

[12] Grudin, Jonathan (2017). *From tool to partner: The evolution of human-computer interaction.* Springer.

[13] Grudin, Jonathan (2023). *ChatGPT and chat history: Challenges for the new wave. IEEE Computer, 56*, 5, 94-100.

[14] Grudin, Jonathan & Jacques, Richard. (2019). Chatbots, humbots, and the quest for artificial general intelligence. *Proc. CHI 2019.* 11 pages.
https://doi.org/10.1145/3290605.3300439

[15] Harwell, Drew (2023). AI-generated child sex images spawn new nightmare for the web. *Washington Post,* June 19.
https://www.washingtonpost.com/technology/2023/06/19/artificial-intelligence-child-sex-abuse-images/

[16] Hashim, Abeerah. (2023). Italy bans ChatGPT over privacy concerns. *PrivacySavvy*, April 2.
https://privacysavvy.com/news/privacy/italy-bans-chatgpt-over-privacy-concerns/

[17] Hill, J., Ford, W.R. & Farreras, I.G. (20 15). Real conversations with artificial intelligence: A comparison between human–human online conversations and human–chatbot conversations. *Computers in Human Behavior, 49*, 245-250. .
https://www.researchgate.net/profile/Jennifer_Hill16/publication/274012711_Real_conversations_with_artificial_intelligence_A_comparison_between_human-human_online_conversations_and_human-chatbot_conversations/links/5a0c4b47a6fdcc39e9bf5c99/Real-conversati

[18] Kerner, Sean Michael. (2023). Einstein AI was good, but Salesforce claims Einstein GPT is even better. *VentureBeat*, March 7.
https://venturebeat.com/ai/einstein-ai-was-good-but-salesforce-claims-einstein-gpt-is-even-better/

[19] Liao, Q.V., Hussain, M.M., Chandar, P., Davis, M., Khazaen, Y., Crasso, M.P., Wang, D., Muller, M., Shami. N.S. & Geyer, W. (2018). All work and no play? Conversations with a question-and-answer chatbot in the wild. *Proc. CHI 2018*, paper 3.

[20] Linn, Allison (2018). Like a phone call: XiaoIce, Microsoft's social chatbot in China, makes breakthrough in natural conversation *The AI Blog*, April 4.
Downloaded June 1 2023. https://blogs.microsoft.com/ai/xiaoice-full-duplex/

[21] Maes, Pattie (1994). Agents that reduce work and information overload. *Comm. of the ACM, 37*, 7, 30-40. https://dl.acm.org/citation.cfm?doid=176789.176792

[22] McCoy, T. (2014). A computer just passed the Turing Test in landmark trial. *Washington Post*, June 9.
https://www.washingtonpost.com/news/morning-mix/wp/2014/06/09/a-computer-just-passed-the-turing-test-in-landmark-trial/

[23] Metz, Cade & Collins, Keith (2018). To give A.I. the gift of gab, Silicon Valley needs to offend you. *New York Times*, Feb. 21. https://www.nytimes.com/interactive/2018/02/21/technology/conversational-bots.html

[24] Microsoft Education Team (2023). Expanding Microsoft Copilot access in education. *Microsoft education blog*, Dec. 14. Accessed 12/23/2024. https://educationblog.microsoft.com/en-us/2023/12/expanding-microsoft-copilot-access-in-education

[25] Mylines (2023). The movie "her" is here. *Reddit,* Dec. 23. Accessed 12/23/2024. https://www.reddit.com/r/ChatGPT/s/XTLVOO5DFr

[26] Nass, Clifford & Yen, Corina. (2010). *The Man Who Lied to His Laptop: What Machines Teach Us About Human Relationships*. Penguin, 2010. ISBN 1-61723-001-4.

[27] NPR (2011). Robot-to-robot chat yields curious conversation. *All things considered,* September 1. https://www.npr.org/2011/09/01/140124824/robot-to-robot-chat-yields-curious-conversation

[28] Park, Mina, Aiken, Milam, & Vanjani, Mahesh. (2018). Evaluating the knowledge of conversational agents. *Southwestern Business Administration Journal, 17*, 1, Article 3. https://digitalscholarship.tsu.edu/cgi/viewcontent.cgi?article=1031&context=sbaj

[29] Porter, Jon. (2021). Alexa's latest celebrity voices are Shaq and Melissa McCarthy. *The Verge*, July 16. https://www.theverge.com/2021/7/16/22579880/alexa-celebrity-voices-shaquille-oneal-melissa-mccarthy

[30] Purtill, James. (2023). Replika users fell in love with their AI chatbot companions. Then they lost them. *ABC Science*, Feb 28. https://www.abc.net.au/news/science/2023-03-01/replika-users-fell-in-love-with-their-ai-chatbot-companion/102028196

[31] Roose, Kevin (2023). Inside the white-hot center of A.I. doomerism. *New York Times*, July 11. https://www.nytimes.com/2023/07/11/technology/anthropic-ai-claude-chatbot.html.

[32] Roose, Kevin (2023). Personalized A.I. agents are here. Is the world ready for them? *New York Times*, November 10. https://www.nytimes.com/2023/11/10/technology/personalized-ai-agents.html?smid=nytcore-android-share.

[33] Salesforce (2023b). Salesforce announces Einstein GPT, the world's first generative AI for CRM. March 7. https://www.salesforce.com/news/press-releases/2023/03/07/einstein-generative-ai/

[34] Schwarz, B. (2023). Microsoft explains how Bing AI Chat uses ChatGPT and Search with Prometheus. Search Engine Land, Feb. 22. https://searchengineland.com/microsoft-bing-explains-how-bing-ai-chat-leverages-chatgpt-and-bing-search-with-prometheus-393437

[35] Shum, H.-Y., He, X., & Li, D. (2018). From Eliza to XiaoIce: challenges and opportunities with social chatbots. *Journal of Zhejiang University Science C*, 19(1), 10-26. https://arxiv.org/abs/1801.01957

[36] Sriram, Akash (2023). Apple blocks update to email app with ChatGPT tech. *Reuters,* March 3. https://www.reuters.com/technology/apple-blocks-update-email-app-with-chatgpt-tech-wsj-2023-03-02/

[37] Sunstein, Cass (2017). *#republic: Divided Democracy in the Age of Social Media.* Princeton University Press.

[38] Ter Stal, Silke, Monique Tabak, Harm op den Akker, Tessa Beinema, & Hermie Hermans. (2020). Who Do You Prefer? The Effect of Age, Gender and Role on Users' First Impressions of Embodied Conversational Agents in eHealth. *International Journal of Human–Computer Interaction, 36*, 9, 881-892. https://www.tandfonline.com/doi/full/10.1080/10447318.2019.1699744

[39] Turing, A. (1949). *London Times* letter to the editor, June 11.

[40] Wakefield, Jane (2023). Would you open up to a chatbot therapist? *BBC News,* April 3. https://www.bbc.com/news/business-65110680

[41] Weizenbaum, J. (1966.) ELIZA -- A computer program for the study of natural language communication between man and machine, *Comm. ACM, 9,* 1, 36-45.

[42] Worswick, Steve (2015). Tweet. Mitsuku has had over 14 million interactions on Kik in just over 2 months. *Twitter*, Nov. 29. https://twitter.com/KukiChatbotDev/status/671095095287029761